\documentclass[11pt]{article}

\usepackage[a4paper]{geometry}
\usepackage{tabularx}
\usepackage{booktabs}
\usepackage{appendix}
\usepackage{bbding}
\usepackage{multirow}
\usepackage{graphicx}
\usepackage{wrapfig}
\usepackage{color, colortbl}
\usepackage{url}
\usepackage{longtable}
\usepackage{lipsum}
\usepackage{xcolor}

\usepackage[leftmargin=20pt,rightmargin=40pt]{quoting}

\usepackage{natbib}
\title{Automated Annotation of Evolving Corpora for Augmenting Longitudinal Network Data: A Framework Integrating Large Language Models and Expert Knowledge\footnote{Corresponding authors are indicated with a dagger ($\dagger$). Please direct correspondence to Xun Pang (email: xpang@pku.edu.cn) or Yansong Feng (emial: fengyansong@pku.edu.cn). }}
\author{Xiao Liu$^{1}$ \hspace{0.1cm} Zirui Wu$^1$ \hspace{0.1cm} Jiayi Li$^1$ \hspace{0.1cm} Zhicheng Shao$^3$\hspace{0.1cm} Xun Pang$^{2,3,\dagger}$\hspace{0.1cm} Yansong Feng$^{1,\dagger}$ \vspace{0.3cm}\\ 
$^1$Wangxuan Institute of Computer Technology, Peking University\\
$^2$School of International Studies and\\ Institute for Carbon Neutrality, Peking University\\
$^3$Analytics Lab for Global Risk Politics, Peking University
}
\date{}
\begin{document}

\maketitle
 \begin{abstract}
Longitudinal network data are essential for analyzing political, economic, and social systems and processes. In political science, these datasets are often generated through human annotation or supervised machine learning applied to evolving corpora. However, as semantic contexts shift over time, inferring dynamic interaction types on emerging issues among a diverse set of entities poses significant challenges, particularly in maintaining timely and consistent annotations. This paper presents the \textit{Expert-Augmented LLM Annotation} (EALA) approach, which leverages Large Language Models (LLMs) in combination with historically annotated data and expert-constructed codebooks to extrapolate and extend datasets into future periods. We evaluate the performance and reliability of EALA using a dataset of climate negotiations \citep{castro2017}. Our findings demonstrate that EALA effectively predicts nuanced interactions between negotiation parties and captures the evolution of topics over time. At the same time, we identify several limitations inherent to LLM-based annotation, highlighting areas for further improvement. Given the wide availability of codebooks and annotated datasets, EALA holds substantial promise for advancing research in political science and beyond.\\

 \end{abstract}


\section{Introduction}
Network data, encompassing information on macro-level social structures and micro-level agent interactions, have become increasingly essential in political analysis of the hyper-connected political world.\footnote{In this paper, we use the term ``network data'' to encompass event data and other relational data.} As these data have accumulated over time and advancements in tracing political systems progress, networks of interest can be monitored longitudinally, giving rise to numerous datasets capturing the evolution of networks over time~\citep{snijders2010introduction,cranmer2017navigating}. Longitudinal network data, with their ability to facilitate quantitative analysis of dynamic relationships and the evolution of systems, have been widely applied to examine important and complex phenomena such as military alliances, political coalitions, international political economies of trade, investment, and value chains, as well as global treaty systems \citep[e.g.,][]{maoz2012, Chaudoin_Milner_Pang_2015, Tsekeris2017, HTWE202034}. These analyses have significantly enhanced our understanding of the patterns and mechanisms underlying political behavior, power dynamics, and systemic interactions. Beyond their importance in substantive studies, longitudinal network data are also increasingly analyzed to advance political methodology. Researchers leverage these data to refine causal discovery and inference techniques, including the development of co-evolution models and causal inference methods that address spillover effects and the interconnected nature of political processes \citep{Snijders2001, Snijders2010, Liu_Pang_2023}.

Longitudinal network data are datasets that capture repeated measurements of a set of entities (nodes) and their defined relationships (ties or edges) across multiple time points. In political science, such datasets are primarily generated from unstructured political texts, such as news reports, official documents, and negotiation summaries. These evolving corpora are annotated mainly by human annotators and increasingly with supervised machine learning models. 
While human annotators often struggle to maintain timely and consistent annotations with massive incoming corpora, supervised machine learning models face challenges on both the \emph{longitudinal} and \emph{network} fronts. The challenging and labor-intensive nature of annotating longitudinal network data has led to significant delays in updating many important datasets in political science. Despite the continuous influx of massive real-time data, these datasets are often not updated promptly. As a result, many studies are forced to rely on significantly outdated datasets, sometimes lagging behind by years or even decades, to test new theories and hypotheses.

In this work, we explore the potential of Large Language Models (LLMs) to annotate longitudinal network data, aiming to overcome the barriers to timely updating such data through high-quality, automated annotation. We focus on addressing two key methodological challenges associated with LLM-based annotation for longitudinal network data.

First, the inclusion of a time dimension requires annotators--whether human or machine--to work on dynamic and evolving semantic contexts to ensure the consistency and coherence of the meaning of the network over time. This is particularly notable in political studies, where concepts are frequently reshaped by shifting semantic representations, especially in studies of emerging, complex, and contentious issues. For example, international climate negotiations not only produce tangible outcomes in climate governance but also continuously redefine the discourse surrounding climate change. A notable instance is the evolution of conflicting relationships initially centered on the divide between developed and developing countries, which has since transitioned into relationships framed by the principle of ``Common but Differentiated Responsibilities (CBDR) and Respective Capabilities (RC)."

Second, the relational nature of network data requires annotators to identify and extract the complex, intertwined relationships among entities. These relationships are often not precisely defined, particularly when they are based on verbal expressions rather than material or physical interactions. The challenge intensifies in diplomatic contexts, where communication is often subtle and laden with implied meanings. In such cases, while some relationships are explicitly stated in the source material, others must be inferred from contextual cues, connotations, or derived through transitivity and existing connections. For example, in multilateral negotiations, if several parties jointly oppose another, both the explicit act of opposition and the implicit alignment or agreement among the opposing countries must be carefully identified and coded.

Recently, there has been growing interest and promising evidence supporting the use of LLMs for annotating data in social science research. Unlike traditional supervised models, LLMs possess advanced language understanding capabilities. These capabilities enable them to interpret nuanced meanings, recognize implicit relationships, and adapt to evolving semantic contexts, making them exceptionally suited for complex and dynamic datasets. This ability to ``read between the lines" positions LLMs as a valuable tool for social science annotations that require contextual interpretation, such as identifying shifts in discourse, detecting subtle changes in relational dynamics, and maintaining consistency across longitudinal datasets. By leveraging these strengths, LLMs could address challenges such as annotating data in politically charged or diplomatically nuanced scenarios, where meanings often transcend literal text, require deeper contextual comprehension, and adapt to evolving political courtesies and social norms. 

However, the potential of LLMs in this context remains largely untapped. Current efforts are predominantly focused on classification tasks involving textual data~\citep{halterman2024codebook,ziems2024can}, such as predicting the sentiment of social media posts. In contrast, our work seeks to extend these applications by exploring how LLMs can be effectively utilized for the more complex task of annotating data that represent dynamic systems with intricate internal structures and interactions.

This research proposes an Expert-Augmented LLM Annotation approach (EALA), which leverages the inherent knowledge and capabilities of LLMs while enabling them to learn from both abstract rules and concrete examples. Specifically, EALA incorporates two key forms of expert knowledge to facilitate the automatic annotation of longitudinal network datasets: an expert-designed codebook and human-annotated historical data derived from applying the codebook. 

A codebook typically provides detailed definitions for nuanced relationship labels, offering clear guidelines for consistent annotation. For instance, the international climate negotiation dataset developed by \citet{castro2017} relies on a comprehensive codebook that defines various types of interactions and specifies coding rules for human annotators. The codebook defines labels such as ``agreement" and ``support", while also providing contextual rules, such as ``agreement should be coded among countries when they collectively oppose another" and ``support is used when the text explicitly states that country 2 (or its statement) was supported by country 1, even when this support is expressed across different sentences". By incorporating such codebooks into prompts, LLMs can engage in deductive learning, applying the predefined rules to generate accurate annotations.
Although labels and coding rules in codebooks can be clearly and thoroughly defined, they are often conceptual in nature and tend to remain abstract. Applying these labels and rules in practice requires contextual interpretation and careful discretion. Even human annotators need extensive training, which is often a continuous and iterative process, especially when the annotation task involves complexities beyond simple classification. To help LLMs better understand the codebook and appropriately apply the labels and rules, EALA utilizes human-annotated historical data. While not capturing more recent developments in the corpus, these data provide a valuable repository of labeled examples that illustrate how various labels have been applied across different contexts, following the guidance of the codebook. By exposing LLMs to this annotated data, either through fine-tuning or in-context learning, the models can learn inductively from concrete examples, recognizing patterns and subtle distinctions in labeling decisions. In other words, we consider both the deductive and inductive learning capabilities of LLMs indispensable and integrate them within EALA to enhance precision and adaptability in annotating complex, evolving corpora.

EALA addresses annotation tasks that go well beyond simple classifications. It is specifically designed to extract and quantify dynamic quadruplets comprising senders, receivers, the connections between them, and the attributes of these connections. Attempting to handle all these subtasks simultaneously often confuses LLMs, leading to errors and hallucinations. To mitigate this, a key component of the EALA framework involves decomposing the overall task into smaller, more manageable subtasks, thereby improving annotation performance. 

We evaluate the performance of EALA across both data-rich and data-scarce settings, using open-source and advanced closed-source LLMs as backbones. Our findings indicate that EALA significantly outperforms traditional supervised models and achieves performance comparable to human annotators. Ablation experiments highlight the contributions of its three key components. Among these, instruction tuning with extensive annotations provides the most substantial improvement, while leveraging annotation examples proves beneficial in data-scarce scenarios. The codebook component offers marginal gains in data-rich settings but becomes crucial when data is limited. Additionally, decomposing the overall task into subtasks, as expected, is essential for maintaining robust performance.

We utilize EALA to annotate relational data among parties engaged in international climate negotiations, encompassing hundreds of countries and negotiation groups over a 30-year span. Political scientists are particularly interested in understanding the complex interactions among states on evolving climate issues and examining the formation of negotiation coalitions that shape the global governance regime for climate change. Much of this research relies on the corpus provided by the Earth Negotiations Bulletin (ENB). Building on an existing codebook and a human-annotated dataset \citep{castro2017}, EALA automatically annotated a negotiation network comprising 45,051 interactions among COP parties across 38 topics from 1995 to the most recent meeting. This application demonstrates how EALA can assist political scientists and climate researchers in analyzing the structure and dynamics of global climate governance, enabling the study of historical trends while keeping pace with the most recent developments.

\section{Motivating Example}
Political scientists study negotiations as dynamic processes aimed at achieving cooperation on complex issues. An important perspective is to conceptualize negotiations as practices of socialization, where new relationships, identities, and perceptions emerge-fostering more sustainable and normatively desirable forms of cooperation \citep{Risse_2000, Comstock_2022}. To analyze the evolving positions of individual parties and the broader negotiation dynamics, structured longitudinal network data are essential. Such data enable empirical investigations through social network analysis and other statistical modeling, offering quantitative and systematic evidence to test theoretical hypotheses about the formation of negotiation networks and the influence these networks exert.

Unlike many international negotiations that often obscure on-site interactions, climate negotiations under the United Nations Framework Convention on Climate Change (UNFCCC) are characterized by high transparency. The proceedings are meticulously documented as daily reports published by the Earth Negotiations Bulletin (ENB), providing public access to summaries of parties' interactions during the meetings of the Conference of the Parties and its subsidiary bodies. However, these negotiations are exceptionally complex, taking place across nearly 100 conferences, involving over 200 parties comprising nation-states and their coalitions, and addressing a wide array of topics. 

ENB monitors sustainable development negotiations and related activities worldwide, publishing daily reports of UNFCCC negotiations since 1995, along with comprehensive sessional reports. These reports detail interactions among negotiation parties, the issues discussed, and the outcomes of the meetings. One notable feature of the ENB is its accessibility and timeliness, as it makes detailed reports publicly available shortly after negotiations conclude.\footnote{https://enb.iisd.org/negotiations/un-framework-convention-climate-change-unfccc} The ENB reports, known for their comprehensiveness and timeliness, have become an increasingly vital source of information for research on negotiations and climate governance \citep{Castro_2014, Vanhala_2016, Weiler2020, Petri2020, Bernardo_2021, Castro_2021, Lefstad_2023}.

Most existing studies utilize information from the ENB summary corpus to address specific research questions within individual projects. To our knowledge, only \citet{castro2017} have undertaken the systematic structuring of this text to support a broader range of political studies on international climate negotiations. Her work involved creating a comprehensive 23-page codebook that encapsulates expert insights on how to transform unstructured text into a quantified longitudinal network. This codebook includes definitions of nodes (nation-states and coalitions) and various types of ties (verbal interactions), detailed categorizations of negotiation topics, definitions of other variables, discussions of ambiguous cases and nuanced distinctions between labels, and coding procedures and rules. Notably, this codebook exemplifies a typical expert-constructed framework in political science, when designed to generate tabular data from textual sources for general applications rather than being tailored to a single research project.

\citet{castro2017} recruited four human annotators to construct a structured longitudinal network dataset covering 461 negotiation days from 1995 to 2013. To validate the codebook, train the annotators, and evaluate inter-coder reliability, the annotators independently coded a randomly selected subset of reports prior to the formal annotation process. Inter-coder agreement was measured using Cohen’s kappa, which revealed that subtle distinctions between certain relationship categories--such as ``Agreement" versus ``With" and ``Opposition" versus ``Criticism"--posed significant challenges, even for human coders.

The annotated longitudinal network dataset of international climate negotiations was publicly released in June 2017. However, no efforts have been made to update the dataset since 
then, leaving it outdated by a decade. This situation is not uncommon in political science, as updating a general-purpose dataset in a timely manner poses significant challenges, especially for researchers lacking a large and stable team dedicated to such data projects. The difficulties are multifaceted, including the complexity of annotating intricate data, the high labor costs associated with human annotation, and the potential inconsistency that arises when transitioning to new annotators during updating efforts. 

In this specific case, the outdatedness is particularly regrettable given the near real-time availability of ENB summaries and the increased prominence of global climate governance over the past ten years. The absence of updated data creates a substantial obstacle for researchers, hindering their ability to generate timely and relevant insights into the evolving dynamics of climate negotiations.

The case of international climate change exemplifies a common data challenge in political science and the broader social sciences, while also showcasing the potential of LLMs to address these issues. Quantitative social science research increasingly depends on annotated data derived from textual corpora, with a growing emphasis on investigating complex social phenomena embedded within rich and nuanced social contexts. By leveraging diverse and real-time information sources, social scientists have gained greater flexibility in accessing and utilizing data. However, this also presents significant challenges in structuring and analyzing data that encompass complex relationships spanning multiple dimensions and levels. The task of data annotation has become increasingly burdensome and technically complex, growing exponentially in scope and difficulty. 

The emergence of powerful tools like LLMs offers promising opportunities for assistance. While LLMs have notable limitations and may initially struggle with complex tasks, the availability of two forms of expert knowledge--systematic annotation guidelines and historical annotations--can be effectively utilized to unlock their full potential \citep{halterman2024codebook}, enabling them to tackle the demanding and labor-intensive tasks of automatically annotating and updating datasets with rich and intricate structures, such as longitudinal network data.


\section{EALA: Task and Method}
\begin{figure}
    \centering
    \includegraphics[width=\linewidth]{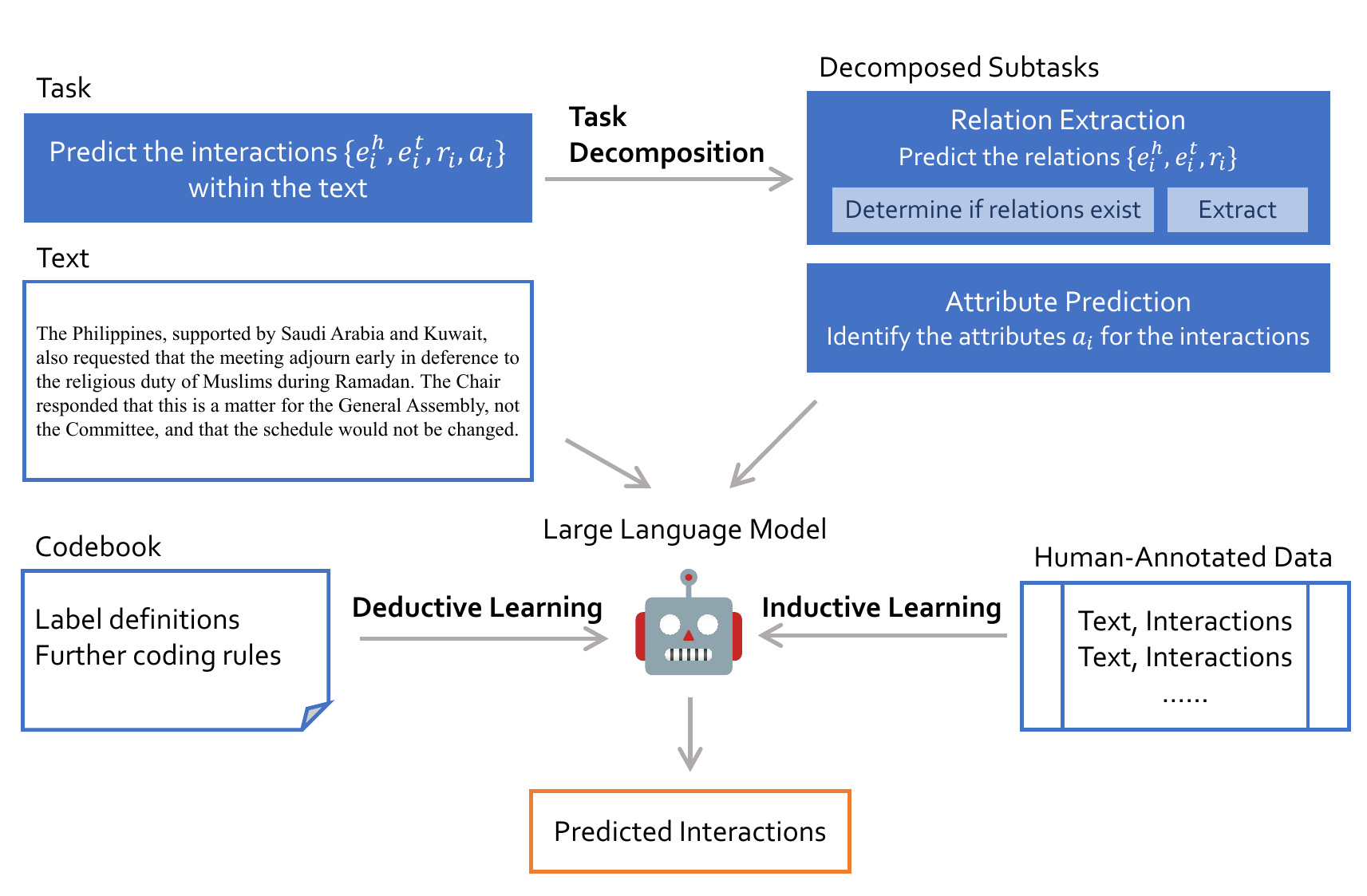}
    \caption{Overview of the EALA framework.}
    \label{fig-framework}
\end{figure}

In this section, we introduce the EALA framework as shown in Figure~\ref{fig-framework}, outlining how we integrate the codebook and annotated data within it. We start by defining the longitudinal network data annotation task. Then we discuss how we calibrate the EALA framework to suit various scenarios and explain the strategies employed to enhance annotations by leveraging the inductive biases of human experts through task decomposition.

\paragraph{Task Formulation}
Given a piece of text $X$, the task is to predict a set of interactions between entities involved in the text, denoted as $Y = \{y_1,  y_2, \ldots, y_n\}$. Each interaction is a quadruplet, $y_i =\{e^h_i, \ e^t_i,\ r_i, \ a_i \}$ for $i =1, 2, \ldots, n$, indicating that it comprises a head entity $e^h_i \in E$, a tail entity $e^t_i \in E$, a relation type $r_i \in R$, and a set of attributes $a_i = \{a^1_i, a^2_i, \ldots, a^m_i\}$, where $a^j_i \in A^j$. The sets of $E$ and $R$ are the predefined space of entities and relation types, and $A^j$ is the label space of an attribute that can change over time.

Consider the motivating example, where \( X \) denotes a paragraph from a daily negotiation report. The set of entities, \( E \), consists of 222 pre-defined COP party countries and the negotiation coalitions they form. Each entity, whether a country or a coalition, can act as the sender (\( e^h \)) or the recipient (\( e^t \)) of an interaction. \citet{castro2017} defined two sets of relations for this dataset; here, we focus on the second set, and \( R \) contains five types of interactions, \( r = 1, 2, \ldots, 5 \), denoting ``On behalf of", ``Support", ``Agreement", ``Delaying proposal", and ``Opposition", respectively. Regarding \( A \), the attributes of interactions, the Castro dataset contains one attribute $A^1$, which contains 19 negotiation topics with each \( a^1_i \) assigned a value from this set. While the topics in her dataset are pre-defined, we aim to make it dynamically adapt to incorporate new topics emerging from negotiations.

The annotation task involves labeling the interactions \( Y_t \) based on the continuously incoming negotiation reports \( X_t \), using the codebook and the historically annotated dataset $\{\mathbf{X}_{t^0}, \mathbf{Y}_{t^0}\}$, where $t^0$ indicates the period before $t$. For simplicity, we will omit the time index in the rest of this paper. Note that since EALA takes a paragraph as a unit of annotation, there will be multiple \( X_t \)s. 

To illustrate the task, consider the following paragraph from the report of the New York Conference on February 7, 1995: 
\begin{quoting}
\texttt{The Philippines, supported by Saudi Arabia and Kuwait, also requested that the meeting adjourn early in deference to the religious duty of Muslims during Ramadan. The Chair responded that this is a matter for the General Assembly, not the Committee, and that the schedule would not be changed.}
\end{quoting}
The annotation task involves annotating all quadruplets \(y=(e^h, e^t, r, a)\) referenced in the text. For instance, one interaction $y_i$ contains $e^h_i=$``Saudi Arabia",  $r_i=$``support",  $e^t_i=$``the Philippines" on a topic $a_i=$``organizational matters". In this paragraph, there is another interaction between Saudi Arabia and Kuwait, which is neither explicitly articulated with a verb nor clearly labeled with a type (e.g., ``agree"). Also, while ``The Chair" is involved in interactions by responding to requests raised by country delegates, these interactions should not be annotated since ``The Chair" does not represent any nation-state or their coalition and falls outside the predefined label space of entities. 

This example illustrates that the task requires the annotator to infer: 1) what the pre-defined negotiation entities referenced in this paragraph are; 2) whether an interaction occurs between two entities; 3) the type of interaction; 4) the directionality of the interaction, i.e., whether it is directed or undirected (mutual); and 5) the issue or topic that the interaction addresses. This task goes beyond simple text extraction and classification, requiring nuanced interpretation and contextual understanding to accurately label the interactions and their attributes.


\paragraph{Utilization of the Codebook} A typical codebook constructed by political scientists for data annotation consists of two main components:
\begin{itemize}
    \item Label definitions. This provides a description of the label space and detailed explanations for each label, including guidelines for when a relation or attribute type should or should not be encoded.
    \item Further coding rules. This includes the general rules that are not limited to a specific label type, such as the rules related to transitivity and derivation of existing interactions.
\end{itemize}
\begin{figure}[!ht]
    \centering
    \begin{tabular}{c}
    \toprule
      \includegraphics[width=1\textwidth]{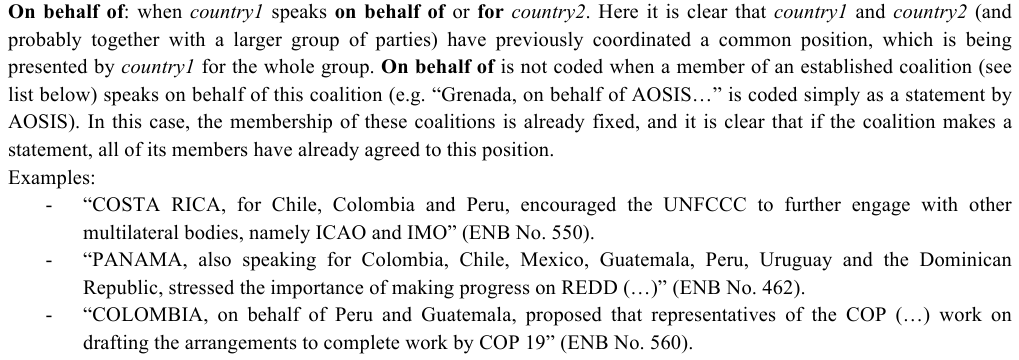} \\
      (a) Label definitions of relations\\
        \midrule
      \includegraphics[width=1\textwidth]{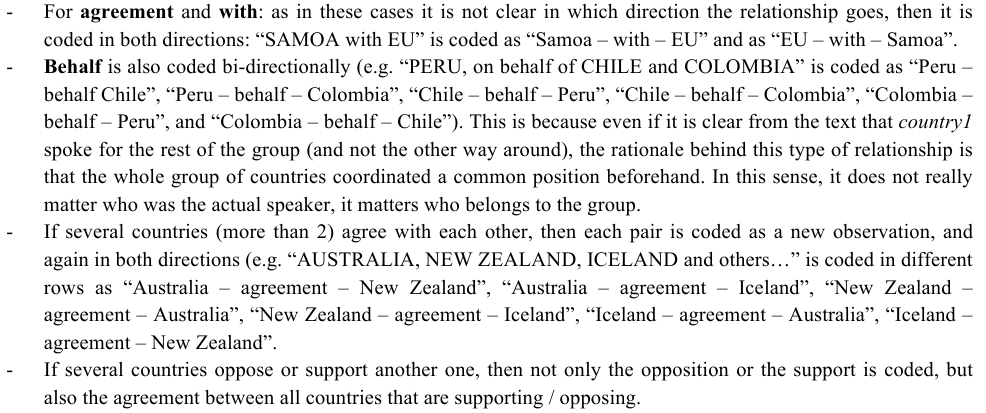} \\
      (b) Further coding rules of relations\\
      \bottomrule
    \end{tabular}
    \caption{Label definitions and coding rules in the codebook (illustrated with the climate negotiation example).}
    \label{codebook}
\end{figure}
In the climate negotiation example, the two components are clearly stated in the codebook but in different sections, as illustrated in Figure~\ref{codebook}.
Both components are incorporated into the EALA prompt, as illustrated in Table~\ref{table-relation-prompt}. These contents are manually retrieved from the original codebook, as the full content of the original codebook is very long and contains information not directly relevant to the annotation task. We leave the automated retrieval of relevant codebook information to future work.

\begin{table*}[!ht]
    \centering
    \small
    \renewcommand{\arraystretch}{1.2}
    \begin{tabularx}{\textwidth}{lX}
    \toprule
    \textbf{Task Instruction} & Please extract all the interactions between parties from this paragraph. \\
    \textbf{Label Definitions} & Each party should be a country or a coalition, in the list ["AILAC", "ALBA", "AOSIS", "Afghanistan", "African Group", "Albania", "Algeria", "Angola", ... (other entities omitted for space)] \\
    & The variable 'interaction' can take the following values. \\
    & On behalf of: when country1 speaks on behalf of or for country2. Here it is clear that country1 and country2 (and probably together with a larger group of parties) have previously coordinated a common position, which is being presented by country1 for the whole group. On behalf of is not coded when a member of an established coalition (see list below) speaks on behalf of this coalition (e.g. “Grenada, on behalf of AOSIS...” is coded simply as a statement by AOSIS). In this case, the membership of these coalitions is already fixed, and it is clear that if the coalition makes a statement, all of its members have already agreed to this position. \\
    & (other relation types omitted for space) \\
    \textbf{Coding Rules} & Further coding rules: \\
    & - If several countries (more than 2) agree with each other, then each pair is coded as a new observation, and again in both directions (e.g. “AUSTRALIA, NEW ZEALAND, ICELAND and others...” is coded as “Australia – agreement – New Zealand”, “Australia – agreement – Iceland”, “New Zealand agreement – Australia”, “New Zealand – agreement – Iceland”, “Iceland – agreement – Australia”, “Iceland agreement – New Zealand”.  \\
    & - If several countries oppose or support another one, then not only the opposition or the support is coded, but also the agreement between all countries that are supporting / opposing. \\
    & (other rules omitted for space) \\
    \textbf{Examples} & [In-context examples in the data-scarce setting] \\
    \textbf{Format Instruction} & The output should be a list of JSON objects. Each JSON object codes one interaction, with keys "Party1", "Party2", and "Relation". \\
    \textbf{Inference Instance} & Paragraph: \\
    & SWITZERLAND, with the EU, called for a “desk review” of non-Annex I communications and, supported by AOSIS, suggested grouping countries with common circumstances for such a review. The PHILIPPINES stressed the need for full-cost financial support. Henriëtte Bersee (Netherlands) and Emily Ojoo-Massawa (Kenya) will consult informally. \\
    & Output: \\
    \bottomrule
    \end{tabularx}
    \caption{Example prompt of EALA in the relation extraction subtask of the climate negotiation dataset.}
    \label{table-relation-prompt}
\end{table*}
\paragraph{Utilization of Human-Annotated Data} Our approach to utilizing annotated data varies depending on the richness of the dataset. 
\begin{itemize}
    \item Data-rich scenario: When conducting continuous annotation based on a publicly available dataset annotated following the codebook, we have sufficient data to support model training. In such cases, we {\it instruction-tune} LLMs with the existing annotated data, i.e., further train LLMs on \emph{<instruction, output>} pairs in a supervised manner~\citep{zhang2023instruction}, enabling them to learn data patterns and distributions. 
    \item Data-scarce scenario: In cases where previously annotated data is unavailable, or when new attributes are introduced, or the label space is modified, only a limited number of human-annotated examples may be available. To address this, we employ {\it in-context learning} by incorporating these examples directly into the prompt~\citep{brown2020language}. Although this approach cannot exhibit the full picture of annotated data, it helps models to better understand the task and follow the instructions.
\end{itemize}

\paragraph{Task Decomposition}
The task of annotating longitudinal network data involves predicting various concepts of multiple interactions, making it challenging for models to tackle it all at once. To address this, we decompose the task based on the inductive bias of human experts into two sequential subtasks: \emph{relation extraction} and \emph{attribute prediction}. In the relation extraction subtask, models are tasked with predicting sets of triplets $(e^h, e^t, r)$, while in the attribute prediction subtask, models focus on identifying attributes ($a$) for the interactions.

During our preliminary experiments, we observed that LLMs tend to hallucinate non-existent interactions when performing relation extraction. To mitigate this issue, we further break down the subtask of relation extraction into two steps: first, determining whether the text includes interactions, and second, extracting interactions from the filtered text. The decomposition strategy allows models to focus more effectively on a simpler task in each step, with a better understanding of the codebook and annotated data related to the current step.

\section{Experiments}
We conduct experiments to evaluate the performance of EALA on the international climate negotiation corpus. Specifically, we aim to answer the following research questions:
\begin{itemize}
    \item RQ1. How do LLMs perform in longitudinal network data annotation, compared to human annotators and traditional supervised models? 
    \item RQ2. Do LLMs gain advantages from using both the codebook and annotated data?
    \item RQ3. What are the best practices in data-rich and data-scarce scenarios?
\end{itemize}

\subsection{Dataset}
Our experiments utilize the human-annotated dataset created by \citet{castro2017}. The dataset and the corresponding codebook are publicly accessible on Harvard Dataverse. This dataset contains annotations of 62,097 dyadic interactions across 5 types, involving 215 entities (nation-states and negotiation coalitions). Note that the label space of entities provided in the codebook contains 222 entities, but not all the entities are detected to have ever been involved in interactions in her sample period. These interactions are categorized by 19 topics and span 461 negotiation dates, covering the period from the 11th Session of the INC in New York in 1995 to COP19 in Warsaw in December 2013. 

\paragraph{Evolving Corpora Collection} The Castro data was annotated using reports from ENB.\footnote{ENB documents various negotiations related to climate, environment, and sustainable development. The Castro data focuses on negotiations under the UNFCCC framework, including meetings of the annual Conference of the Parties (COP), sessions of the UNFCCC’s permanent subsidiary bodies, ad-hoc negotiation groups, and specific technical workshops.} ENB reports consist of both daily and summary reports: daily reports capture the interactive behaviors of parties on each negotiation day, such as expressions of support, opposition, or speaking on behalf of coalitions, with all recorded interactions being verbal; summary reports provide an overview of the entire meeting, emphasizing key outcomes such as conclusions and decisions adopted by the COP.

ENB updates negotiation reports in a timely manner. To evaluate the performance of EALA, we collect reports from the official ENB website, starting on the same date as the Castro data to the Bonn Climate Change Conference in June 2024.\footnote{https://enb.iisd.org/. The latest summaries available are for COP 30 in December 2024 in Baku.} There are 844 reports in total. Given that most original annotations were based on daily reports, we also annotate interactions within daily reports for consistency. Reports related to frameworks outside the UNFCCC, such as the Intergovernmental Panel on Climate Change (IPCC), were excluded. This amounts to a total of 678 daily reports, with an average of 39 paragraphs and 2,516 words per report (about 64 words per paragraph).

\paragraph{Human-Annotated Historical Dataset} The Castro dataset was created by employing four coders who manually annotated the corpus, following the guidelines outlined in the codebook. The code reports high consistency of relation annotations among coders, with Cohen's Kappa between pairs of coders ranging from 0.90 to 0.98. However, topic annotations exhibited only fair consistency, with Kappa ranging from 0.25 to 0.72, reflecting the complexity of climate negotiations. In terms of label space, interactions were initially coded into seven relation types, later simplified to five by merging similar categories. The final relation space $R$ comprises the types: \texttt{On behalf of}, \texttt{Support}, \texttt{Agreement}, \texttt{Delaying proposal}, and \texttt{Opposition}. Each interaction was also associated with a topic from a predefined set of 19 labels, such as \texttt{Adaptation} and \texttt{LULUCF} (land use, land use change and forestry).


\paragraph{Dynamic Topic Space Construction}
\begin{wrapfigure}{R}{0.5\textwidth}
    \centering
    \includegraphics[width=0.48\textwidth]{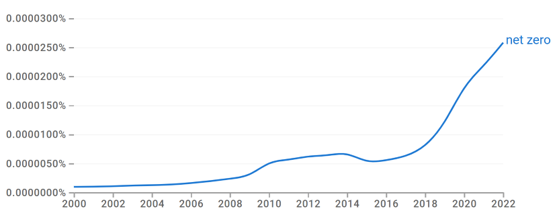}
    \caption{Frequency of the term \emph{net zero} from Google Ngram Viewer.}
    \label{fig-net-zero}
\end{wrapfigure}
The Castro dataset restricted annotations to fixed sets of predefined labels for entities, types of relations, and topics. While this approach works reasonably well for entities and types of relations, as their categories are relatively stable (despite the occasional emergence or dissolution of negotiation groups/coalitions), it falls short when it comes to negotiation topics. For example, the term \emph{net zero} only gained prominence after 2018, as illustrated in Figure~\ref{fig-net-zero}.  Topics are a crucial aspect, serving both as attributes and as outcomes of negotiations. Limiting topics to a fixed set of predefined labels fails to leverage the evolving nature of discourse, thereby missing an important dynamic that holds significant interest for political scientists. To address this, we construct a dynamic topic space that evolves over time to better capture these changes.

\begin{table}[t]
\centering
\small
    \begin{tabular}{lccc}
        \toprule
         & \textbf{Stage 1} & \textbf{Stage 2} & \textbf{Stage 3} \\
        \midrule
        Time Range & 1995 - 2013 & 2014 - 2018 & 2019 - 2024 \\
        Number of Topics Added & 30 & 3 & 3 \\
        Example Topic & Emissions & Non-Market Approaches & Net Zero \\
        Topics Revised by Annotators & \multicolumn{3}{c}{27.8\%} \\
        \bottomrule
    \end{tabular}
    \caption{Statistics and examples of topics captured dynamically.}
    \label{table-topic-space}
\end{table}

We divide the timeline into three stages, as detailed in Table~\ref{table-topic-space}. The base topic space was developed using meetings from 1995 to 2013. We use LLMs to extract topic words from summary reports, cluster them into 30 topics, and provide names and descriptions for each topic. For later stages, additional topic words from summary reports are analyzed to determine if they fall within existing topics or warrant new categories. LLMs generate names and descriptions for new topics accordingly. This process results in a topic space comprising 36 topics, which can be further expanded with future meetings. To ensure the quality of the topics, we ask human annotators to review and revise the topics, and 27.8\% of the topic descriptions are revised. Additional information on topic space construction is provided in Appendix~\ref{appendix-topic-space}.

\paragraph{Evaluation Data}
Our evaluation investigates two subtasks: \emph{relation extraction} and \emph{attribute prediction}. Both tasks are structured around paragraphs.

For relation extraction, we operate under a \emph{data-rich scenario} where models are provided with extensively annotated interactions, accompanied by a codebook that includes definitions of relation types and detailed coding rules. The models receive a paragraph as input and are tasked with predicting all the interactions $(e^h,  e^t, r)$ within the paragraph. We sample 3,600 paragraphs from the annotated reports as training data and sample 256 paragraphs as test data. The reports used for the training data predate those used for the test data.\footnote{Only a portion of the annotations in the Castro dataset include corresponding quotes. Specifically, for issues numbered $\leq$351, about 97\% of annotations have quotes, whereas for issues numbered $>$ 351, only 12\% include quotes. Based on the provided quotes, the search window appears to be limited to a single sentence. To ensure that most annotations could be mapped to paragraphs, we sampled training and testing data from the Castro dataset with issue numbers $\leq$351. Using fuzzy matching (with >90\% word overlap), we mapped the quotes to paragraphs. Out of 25,157 annotations with quotes, 24,795 were successfully matched to a single paragraph. }

Conversely, attribute prediction is designed for a \emph{data-scarce} scenario. Models are supplied only with a few examples and the codebook, which outlines topic definitions. Models are given a paragraph along with the interactions within it, and are instructed to identify the topic of each interaction. 
Due to the absence of existing data for the dynamic topic space, we recruit eight coders to annotate 50 sampled paragraphs containing 394 interactions. The interactions are from the original dataset to ensure the validity. Each interaction is annotated by two coders, resulting in an inter-annotator agreement of 0.79. Disagreements are resolved through discussion, leading to the final annotations. We select one paragraph as an example and use the remaining 49 paragraphs as the test data.\footnote{We provide only a single example paragraph because both the example and the codebook are quite long. Additionally, each example includes 4-6 interactions, allowing models to learn the attribute prediction task for multiple interactions.} This sampling process is repeated three times, and the average performance is reported.

\subsection{Experimental Setups}
\paragraph{Models} We select GPT-4o~\citep{hurst2024gpt} and Llama-3-8B~\citep{dubey2024llama} as the backbones of our framework, representing closed-source and open-source models respectively. The specific versions used are \texttt{gpt-4o-2024-08-06} (over 170b parameters) and \texttt{Llama-3-8B-Instruct} (8b parameters). In the relation extraction subtask, we also create a traditional supervised baseline by finetuning a small model T5-base~\citep{raffel2020exploring}. This is a smaller, 220m-parameter model from Google's T5 family and has been commonly used in social science annotation projects~\citep{ziems2024can}.

\paragraph{Implementation Details} During the instruction tuning of Llama-3-8B, we optimize the model using the AdamW optimizer with a learning rate set at 2e-5. The training is conducted over 3 epochs with a batch size of 8. For inference with both GPT-4o and Llama-3-8B, we set the maximum length to 4,096 and the temperature to 0. We finetune T5-base with a learning rate of 5e-5 and a batch size of 4 for 4 epochs, with the hyperparameters selected through grid search.

\paragraph{Evaluation Metrics}
We use precision and recall for relation extraction, and use accuracy and macro F1 for attribute prediction.

\subsection{Evaluation Results}
\begin{table}[t]
\centering
\small
    \begin{tabular}{lcc}
        \toprule
         & \textbf{Precision} & \textbf{Recall} \\
        \midrule
        Individual Human Annotators & 67.5 & 89.2 \\
        Human Annotators with Discussion & 67.2 & \textbf{90.3} \\
        EALA (Llama-3-8B) & \textbf{74.0} & 90.0 \\
        \bottomrule
    \end{tabular}
    \caption{Performance of EALA compared to human annotators in relation extraction on a subset of the test data. All numbers are in percentages (\%).}
    \label{table-relation-human}
\end{table}

\begin{table}[t]
\centering
\small
    \begin{tabular}{lccccc}
        \toprule
        \textbf{Model} & \textbf{Codebook} & \textbf{Annotation} & \textbf{Decomposition}& \textbf{Precision} & \textbf{Recall} \\
        \midrule
        T5 & \XSolidBrush & Full & \Checkmark & 33.2 & 10.1 \\
        \midrule
        \multirow{7}{*}{Llama-3-8B} & \XSolidBrush & \XSolidBrush & \Checkmark & 0.9 & 1.0 \\
        & \Checkmark & \XSolidBrush & \Checkmark & 3.6 & 3.6 \\
        & \XSolidBrush & Examples & \Checkmark & 20.0 & 20.1 \\
        & \Checkmark & Examples & \Checkmark & 27.2 & 26.1 \\
          & \Checkmark & Full & \XSolidBrush & 73.7 & 54.0 \\
        & \XSolidBrush & Full & \Checkmark & 76.2 & 87.9 \\
        \rowcolor[gray]{0.95} & \Checkmark & Full & \Checkmark & \textbf{77.7} & \textbf{90.3} \\
        \midrule
        \multirow{4}{*}{GPT-4o} & \XSolidBrush & \XSolidBrush & \Checkmark & 20.2 & 14.9 \\
        & \Checkmark & \XSolidBrush & \Checkmark & 7.7 & 12.9 \\
        & \XSolidBrush & Examples & \Checkmark & 46.6 & 46.5 \\
        & \Checkmark & Examples & \Checkmark & \textbf{52.3} & \textbf{68.3} \\
        \bottomrule
    \end{tabular}
    \caption{Performance of different models and settings in the task of relation extraction. The EALA framework is shaded in gray. All numbers are in percentages (\%).}
    \label{table-relation-ablation}
\end{table}

\begin{table}[t]
\centering
\small
    \begin{tabular}{lccc}
        \toprule
        \textbf{Model} & \textbf{Bidirectionality} & \textbf{Transitivity} & \textbf{Derivation} \\
        \midrule
        EALA (Llama-3-8B) w/o Codebook & 94.7 & 92.2 & 59.1 \\
        EALA (Llama-3-8B) & \textbf{99.6} & \textbf{96.1} & \textbf{88.2} \\
        \bottomrule
    \end{tabular}
    \caption{Proportion of compliance with rules involving multiple interactions. All numbers are in percentages (\%).}
    \label{table-relation-rule}
\end{table}

\begin{table}[t]
\centering
\small
    \begin{tabular}{lcccc}
        \toprule
        \textbf{Model} & \textbf{Codebook} & \textbf{Annotation} & \textbf{Accuracy} & \textbf{Macro F1} \\
        \midrule
        \multirow{4}{*}{Llama-3-8B} & \XSolidBrush & \XSolidBrush & 12.5 & 14.4 \\
        & \Checkmark & \XSolidBrush & 24.1 & 17.0 \\
        & \XSolidBrush & Examples & 19.6 & 13.8 \\
        \rowcolor[gray]{0.95} & \Checkmark & Examples & \textbf{25.1} & \textbf{22.9} \\
        \midrule
        \multirow{4}{*}{GPT-4o} & \XSolidBrush & \XSolidBrush & 55.1 & 35.2 \\
        & \Checkmark & \XSolidBrush & \textbf{63.8} & 40.4 \\
        & \XSolidBrush & Examples & 53.6 & 32.0 \\
        \rowcolor[gray]{0.95} & \Checkmark & Examples & 63.7 & \textbf{46.4} \\
        \bottomrule
    \end{tabular}
    \caption{Performance of different models and settings in the task of attribute prediction. The EALA framework is shaded in gray. All numbers are in percentages (\%).}
    \label{table-topic-ablation}
\end{table}

\paragraph{Overall Performance of EALA}
We evaluate the performance of EALA in comparison to human annotators and traditional supervised models in the relation extraction subtask, using the Castro dataset as a benchmark.
For human annotation, we recruited ten coders to annotate five daily reports comprising 153 paragraphs sampled from the test set, with each paragraph being annotated by two different coders. We present their individual performance, performance after discussion, and EALA (Llama-3-8B)'s performance on this sampled subset in Table~\ref{table-relation-human}. EALA demonstrates 6\% higher precision and comparable recall compared to human annotators, indicating its effectiveness in annotating longitudinal network data and its potential to assist humans in real annotation tasks.

To determine if LLMs surpass traditional supervised models in annotation, we compare the performance of Llama-3 and T5, as shown in Table~\ref{table-relation-ablation}. Following the traditional way of training supervised models, we only provide annotated data to them. Llama-3 achieves 43\% higher precision and 78\% higher recall than T5 under the same setting, highlighting that, with advanced language understanding and reasoning capabilities, LLMs are much more proficient in executing the annotation task.

\paragraph{Impacts of Different Components}
To evaluate the impact of each component, we conduct ablation studies on EALA. The results for relation extraction are presented in Table~\ref{table-relation-ablation}, and the results for attribute prediction are presented in Table~\ref{table-topic-ablation}. 
\begin{itemize}
    \item \textbf{Codebook.} Incorporating the codebook into the prompt enhances annotation performance across nearly all settings for both tasks. On average, this results in an increase of 1\% in precision and 6\% in recall for relation extraction, and an improvement of 9\% in accuracy and 8\% in F1 score for attribute prediction. 
    
    The improvement is particularly notable when LLMs are not provided with annotated data or are only given annotation examples (with a 7\% increase in precision and a 6\% increase in recall for relation extraction). In these cases, the codebook plays a crucial role by informing models about label definitions. Conversely, when LLMs are trained on the annotated data, they can learn the label definitions directly from the data. However, the codebook still provides value by conveying general coding rules. These rules are often abstract and may involve multiple interactions, making them difficult to derive solely from annotated data. We compare the models' compliance with these rules in Table~\ref{table-relation-rule}, observing that EALA (Llama-3-8B) adheres to all three rules more effectively when assisted by the codebook, compared to the version without it. Detailed descriptions of the rules are available in Appendix~\ref{appendix-rules}.
    
    \item \textbf{Annotated Data.} We compare three variants regarding the use of annotated data: not providing annotations to the model, providing annotation examples within the prompt, and instruction-tuning the model with the complete set of annotations. The last approach is feasible only for data-rich scenarios with open-source models.
    For topic extraction, compared to providing no annotations, including annotation examples benefits models, enhancing precision by 28\% and recall by 32\% for relation extraction, and improving both accuracy and F1 score by 2\% for attribute prediction. The improvements are greater in relation extraction since this task is less familiar and more challenging for LLMs. Without examples, the models struggle even to generate outputs in the desired format, such as producing interactions with only one entity. 

    Instruction-tuning models with the full set of annotations yields greater improvements. Compared to the example-only variant, training on annotated data results in an average increase of 53\% in precision and 66\% in recall, highlighting the advantage of leveraging the large amount of data. With this training, the smaller Llama-3-8B model substantially outperforms the more powerful GPT-4o in annotation performance. For attribute prediction, the relatively low accuracy and F1 scores achieved by EALA, without instruction tuning, are unsurprising given the presence of over 30 labels.
    
    \item \textbf{Task Decomposition.} Due to the complexity of the relation extraction subtask, we use expert experience to break it down into two distinct steps. Compared to the model trained in an end-to-end manner, task decomposition results in an improvement of 36\% in recall, reducing hallucinations (i.e., generating nonexistent interactions) drastically. 
\end{itemize}

\paragraph{Takeaways of Annotating Longitudinal Network Data}
Both the codebook and annotated data prove to be beneficial in annotating longitudinal network data. In data-rich scenarios, instruction-tuning smaller LLMs with extensive annotations may be more effective than relying on more powerful closed-source LLMs. While in data-scarce scenarios, prompting powerful LLMs using in-context learning with annotation examples is suggested. Simultaneously, the codebook provides detailed label definitions and coding rules, which are crucial for maintaining robust performance and complement the annotated data effectively. Additionally, breaking down complex tasks into manageable steps is worth considering. While LLMs are powerful tools, the incorporation of expert insights is crucial to maximize their effectiveness.


\section{Empirical Evaluation of EALA-Annotated Climate Negotiation Dataset}
In addition to evaluations based on quantitative metrics, we further assess the performance of the proposed EALA framework through empirical analysis of the annotated data. However, due to the lack of detailed information regarding the search windows used by \citet{castro2017}, a direct comparison based on matching annotated $y_i$s between the two datasets is not feasible. Notably, varying search windows can yield significantly different $y_i$ annotations, as responses to actions during negotiations may appear multiple paragraphs later. Consequently, we focus on comparing the general patterns of the datasets through statistical features rather than observation-by-observation comparisons.

\begin{figure}[ht]
    \centering
    \begin{tabular}{cc}      
    \includegraphics[width=0.45\textwidth]{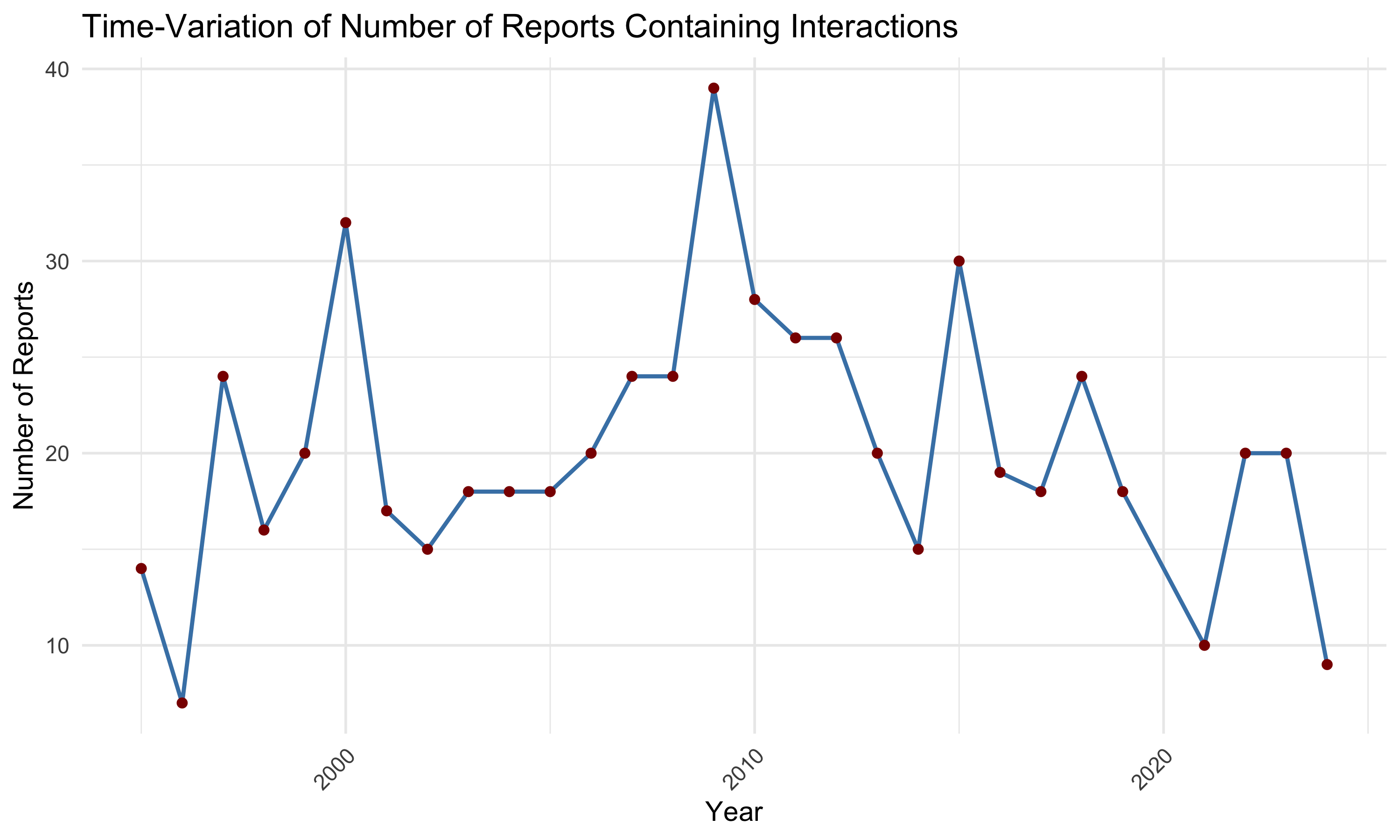} &         \includegraphics[width=0.45\textwidth]{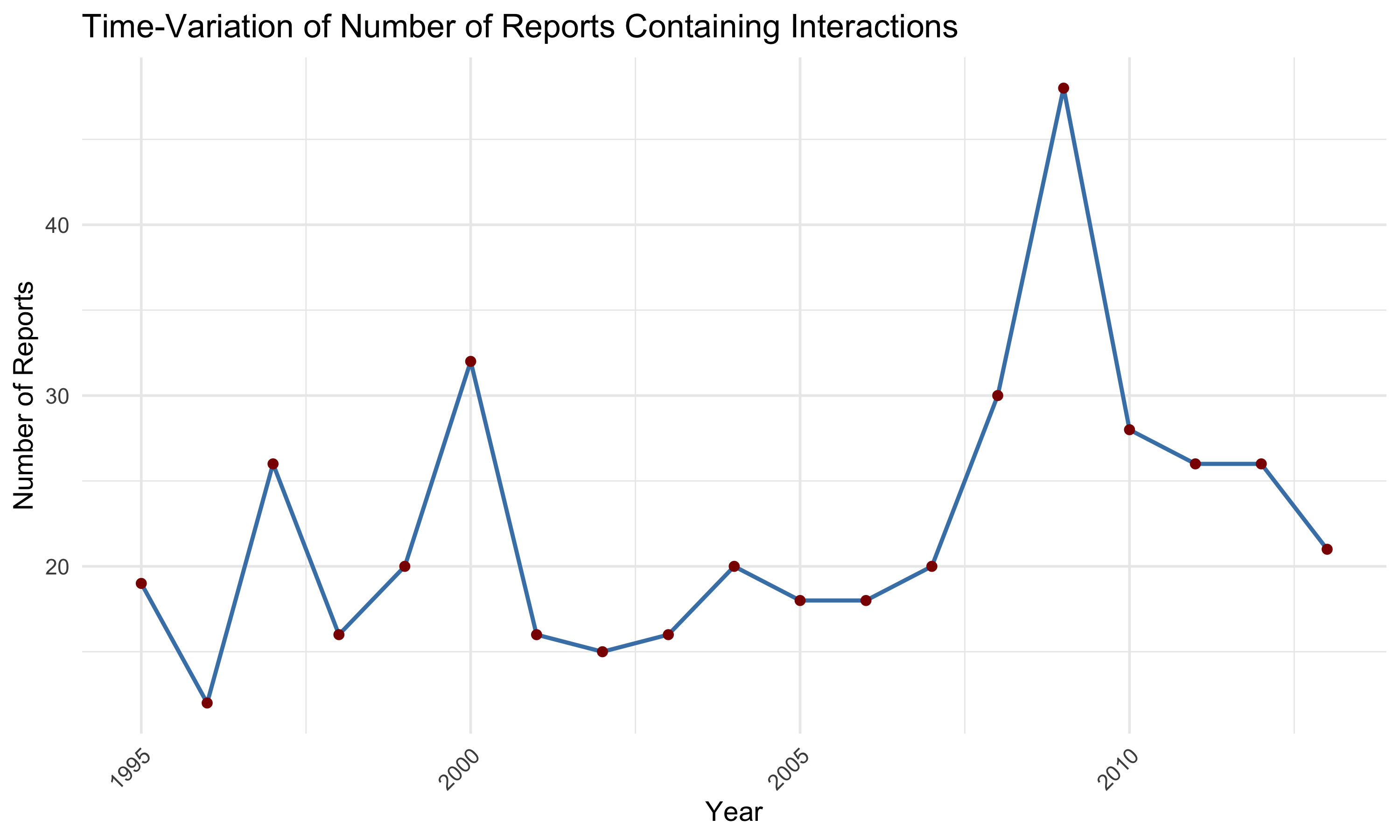} \\
    \includegraphics[width=0.45\textwidth]{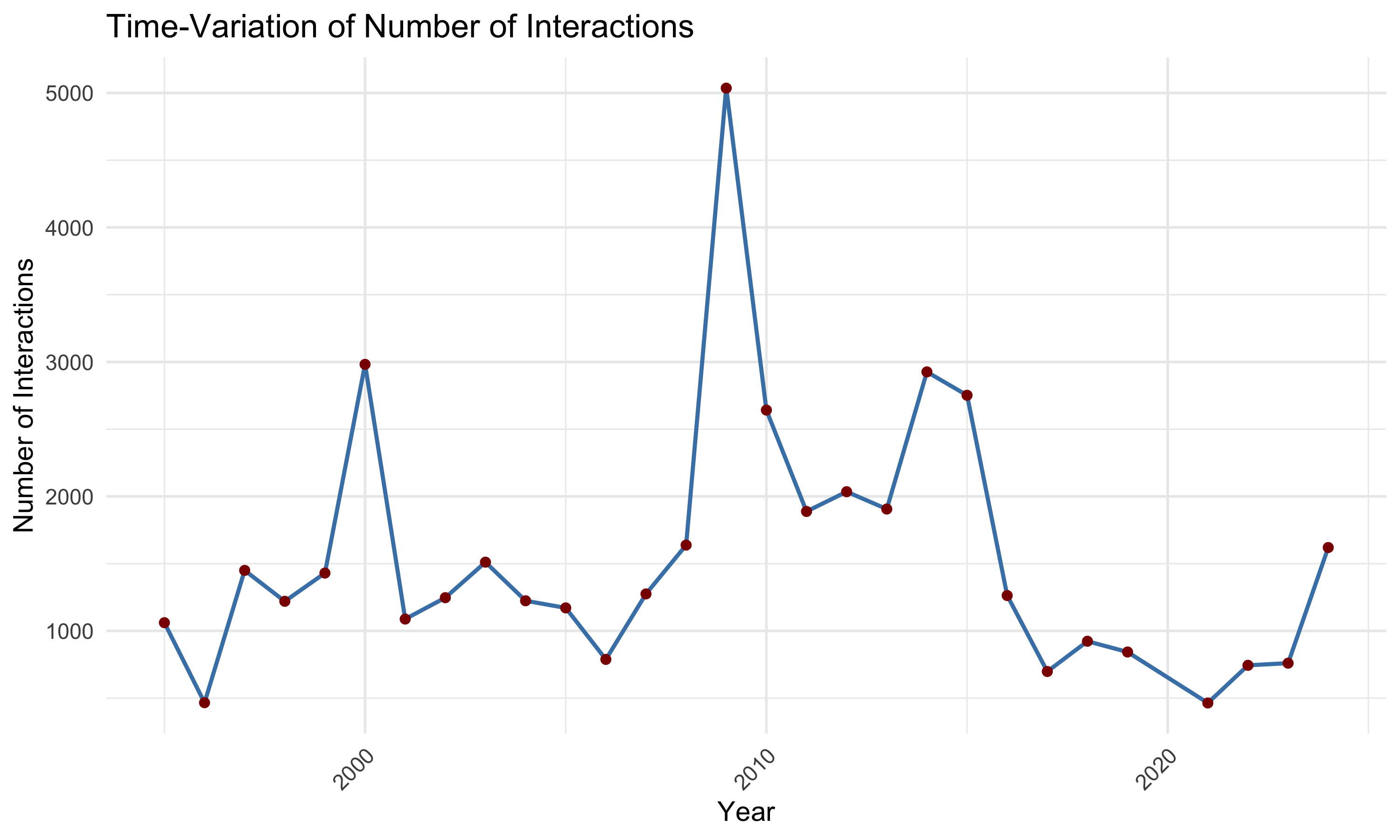} &
    \includegraphics[width=0.45\textwidth]{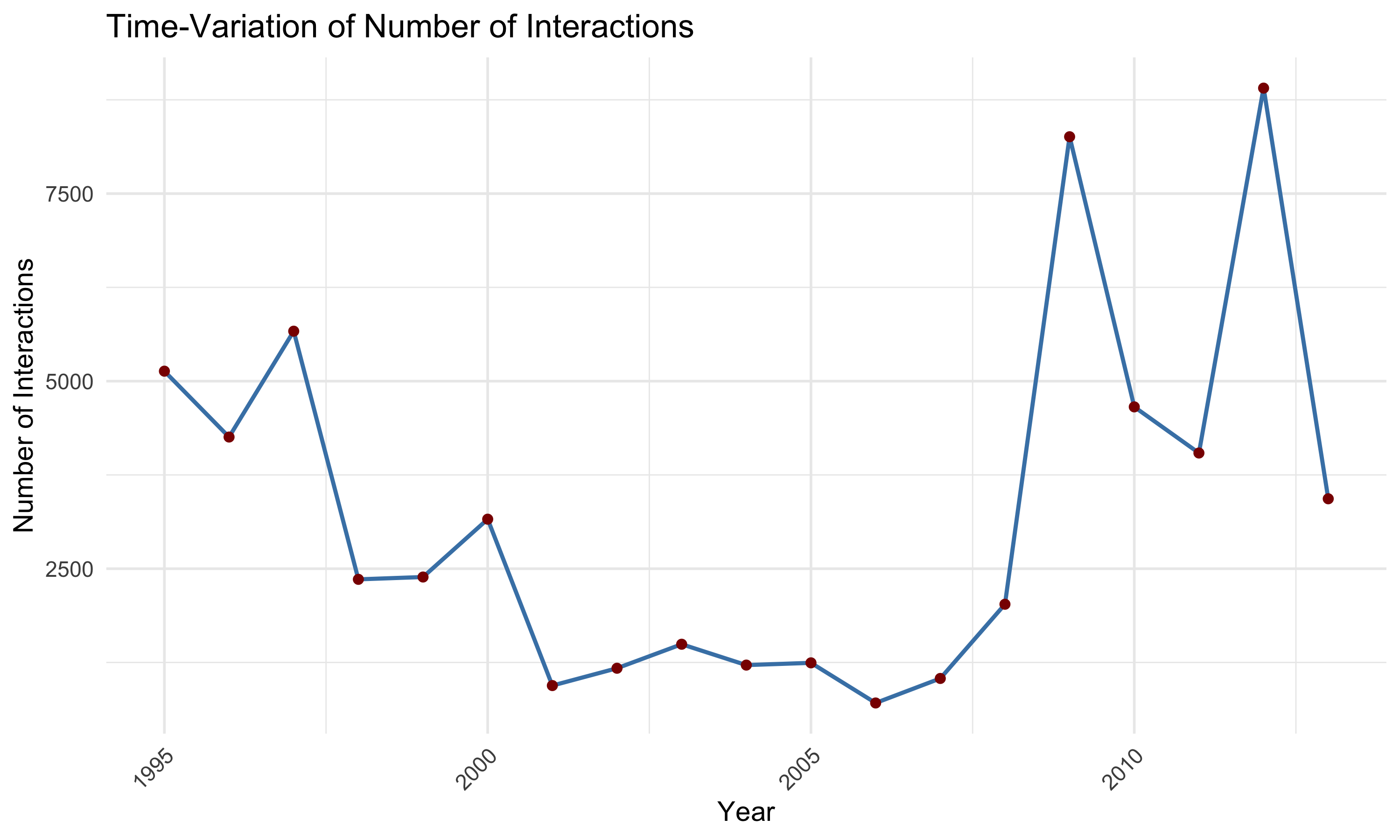} \\
    (a) EALA annotation (1995-2024) & (b) Castro annotation (1995-2013)\\
    \end{tabular}
    \caption{Yearly distribution of reports and interactions.}
    \label{textNum}
\end{figure}

EALA identified 589 out of 678 daily reports as containing interactions. The number of daily reports fluctuated over the 30 sample years, as illustrated in the upper panel of Figure~\ref{textNum}(a). From the 589 reports, EALA annotated 45,051 interactions in total. The fluctuation in the number of interactions closely mirrors the variation in the number of reports in a year, with a correlation of 0.724, as illustrated in the lower panel of Figure~\ref{textNum}(a). Both the number of reports and interactions peaked in 2009, the year of the Copenhagen Climate Summit, before declining sharply after 2015, the year of the Paris Agreement. The post-Paris Agreement period exhibits significantly sparser interactions under the UNFCCC framework, compared to the years preceding it.

When comparing the number of LLM-detected reports containing interactions to the number of reports in the Castro dataset (as shown in the first row of Figure~\ref{textNum}), their yearly distributions are broadly similar, though the LLM-detected numbers are slightly lower than those in the Castro data. However, the annotated numbers of interactions show a stark contrast between the two methods, as illustrated in the second row of Figure~\ref{textNum}. 







In some cases, the gap is too large to be attributed to randomness. For instance, in 1995, the first sample year, the Castro dataset recorded 5,133 interactions, while the EALA-generated data annotated only 1,061 interactions. Even accounting for the difference in the number of reports, this gap remains substantial. Further examination of the Castro dataset revealed EALA's omission of some indirect interactions. But more importantly, in the Castro dataset, numerous annotated interactions are not found in the provided corresponding quotes, and in many cases, even the entities involved in these interactions are absent from the quoted text. This inconsistency indicates that the dataset does not reliably link annotations to specific quotes, and the search window for interactions extends far beyond the provided sentences. In contrast, EALA precisely uses a paragraph as its search window. The mismatched annotations and quotes in the Castro dataset also diminish the effectiveness of instruction tuning in EALA. This analysis highlights that even human-coded datasets for complex annotation tasks are prone to errors, arbitrariness, and subjectivity. Furthermore, EALA is expected to achieve even greater performance when instruction tuning is supported by high-quality human-annotated datasets with reliable and well-defined annotation-to-text mappings.

Next, we examine the annotated distributions for each of the four elements in the quadruplet \( y_i \). 

\begin{figure}[ht]
    \centering
    \begin{tabular}{cc}
            \includegraphics[width=0.47\textwidth]{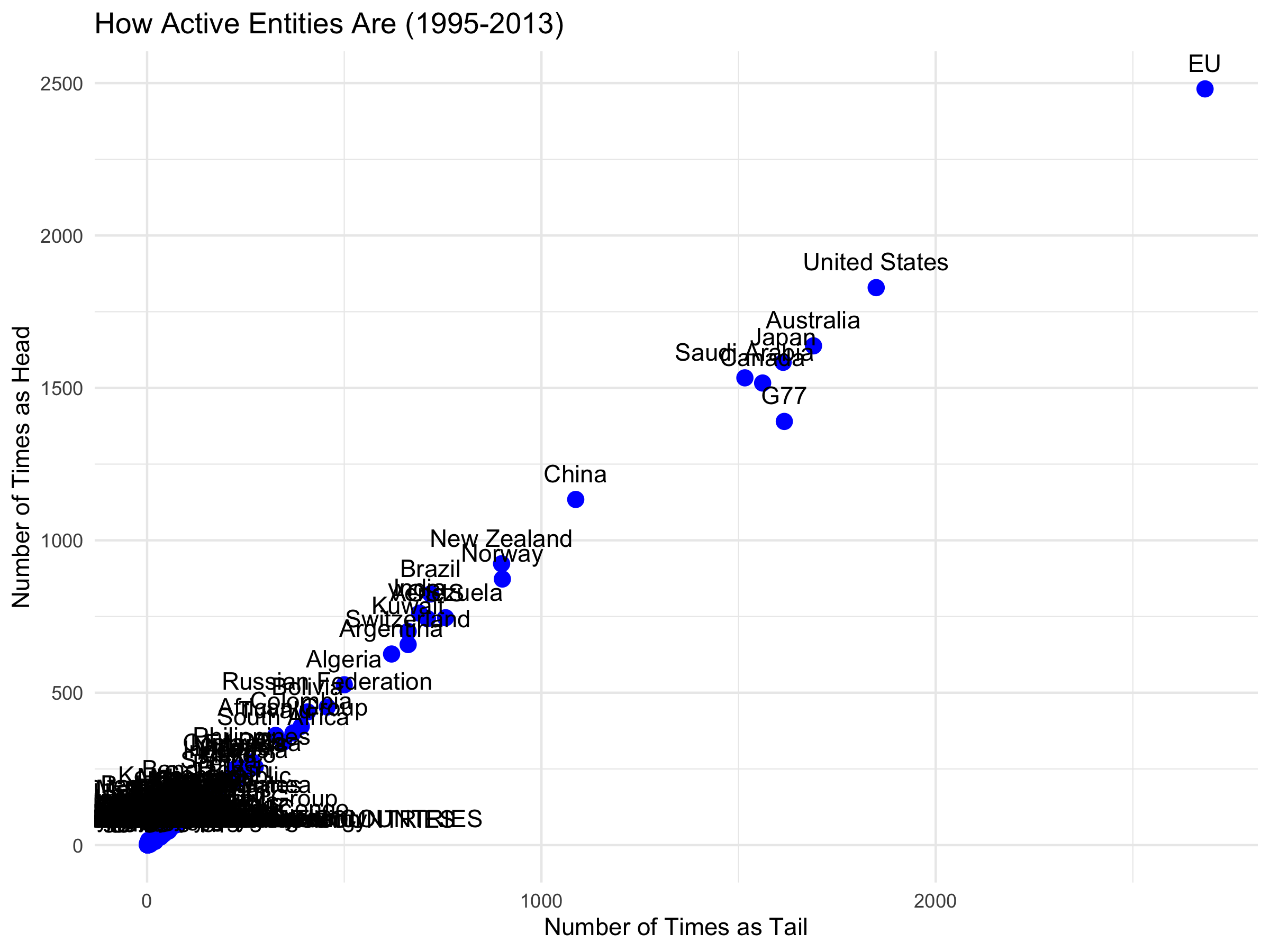}&
       \includegraphics[width=0.47\textwidth]{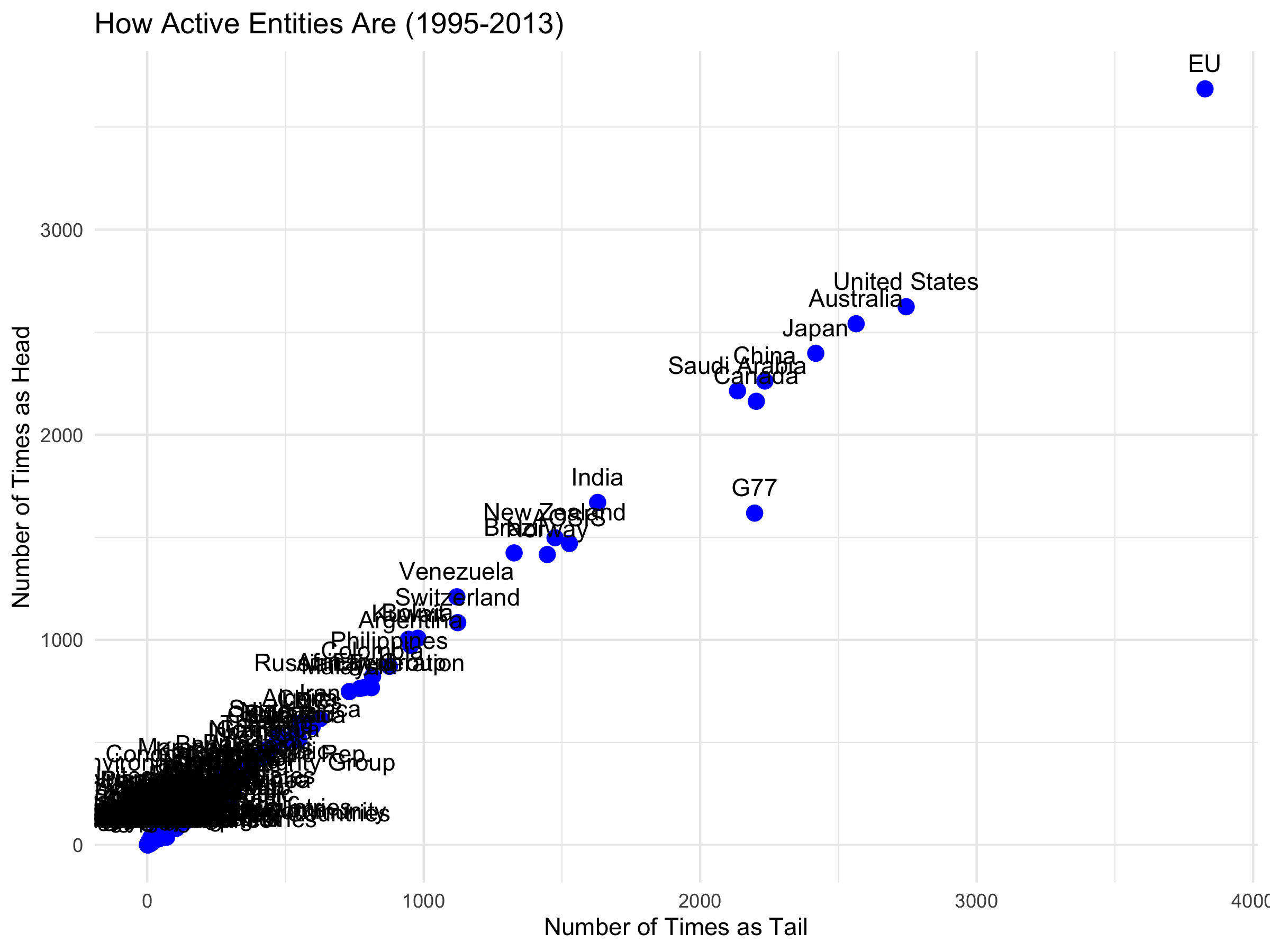}\\
       (a) EALA annotation (1995-2013) & (b) Castro annotation (1995-2013)
    \end{tabular}
    \caption{Degree of activity of negotiation entities.}
    \label{active}
\end{figure}

\paragraph{Negotiation Entities}  Figure~\ref{active} compares the degree of activity of negotiation entities between the EALA-annotated dataset and the Castro dataset. The comparison is conducted for the subsample spanning 1995 to 2013, and Figure~\ref{active2} in the appendix depicts the full period from 1995 to 2024. As illustrated in panels (a) and (b), the machine- and human-annotated datasets largely align in identifying the most active entities and highlight the hierarchical nature of the network. Specifically, while the majority of entities rarely engaged in negotiations (or at least their engagement was not significant enough to be recorded in the reports), a small subset of entities consistently exhibited exceptionally high levels of activity. In addition, both datasets reveal that entities that are highly active as senders are also more frequently recipients of actions, indicating that negotiations mostly occur within the small subset of active entities. 
However, the LLM annotation of negotiation entities demonstrates several noteworthy points, underscoring the need for future research and solutions: 

{\it Detection Beyond the Pre-defined Scope.} LLMs identified entities beyond nation-states and negotiation coalitions, including supra-national and sub-national organizations (e.g., the Technology Executive Committee, LULUCF, The Saudi Central Bank, and the Sahel Observatory), individuals (e.g., Tara Shine, Tosi Mpanu-Mpanu, and Muyungi), sub-national jurisdictions (e.g., Sharm el-Sheikh, Nairobi), and private entities (e.g., Munich Climate Initiative). While these detections show the breadth of LLMs' capabilities, they also include irrelevant entities that require manual correction.  

{\it Emerging Entities.}   LLMs successfully detected emerging negotiation groups and entities not included in pre-defined space, such as the Visegrad Group and SIDS (Small Island Developing States). This showcases LLMs' ability to identify new patterns and entities in dynamic contexts.  

{\it Erroneous Entity Detection.} LLMs sometimes misidentified non-actor nouns as entities, such as ``Sinks'', ``Sahara'', ``Women'', ``Women and Gender'', and ``Civil Society''.  LLMs also generated strange artifacts, such as ``On an option for a framework'', ``Facilitator La Vi$\backslash$x96a'', and ``T$\backslash$x9frkiye'', reflecting challenges in text interpretation and accuracy.  

Therefore, LLMs detected 401 entities in total, significantly more than the 222 entities in the Castro dataset. While most of Castro's pre-defined entities were included in the LLM-detected set, this over-detection indicates the model's tendency to go outside the label space. Many of the errors or over-detections can only be reliably addressed through manual verification and post-processing, emphasizing the current limitations of LLM-based annotation.  

These findings highlight the need for further exploration into refining LLM models, improving their contextual understanding, and developing robust post-processing techniques to enhance the accuracy and reliability of annotated datasets.

\begin{figure}[ht]
    \centering
    \begin{tabular}{c}
      \includegraphics[width=0.9\textwidth]{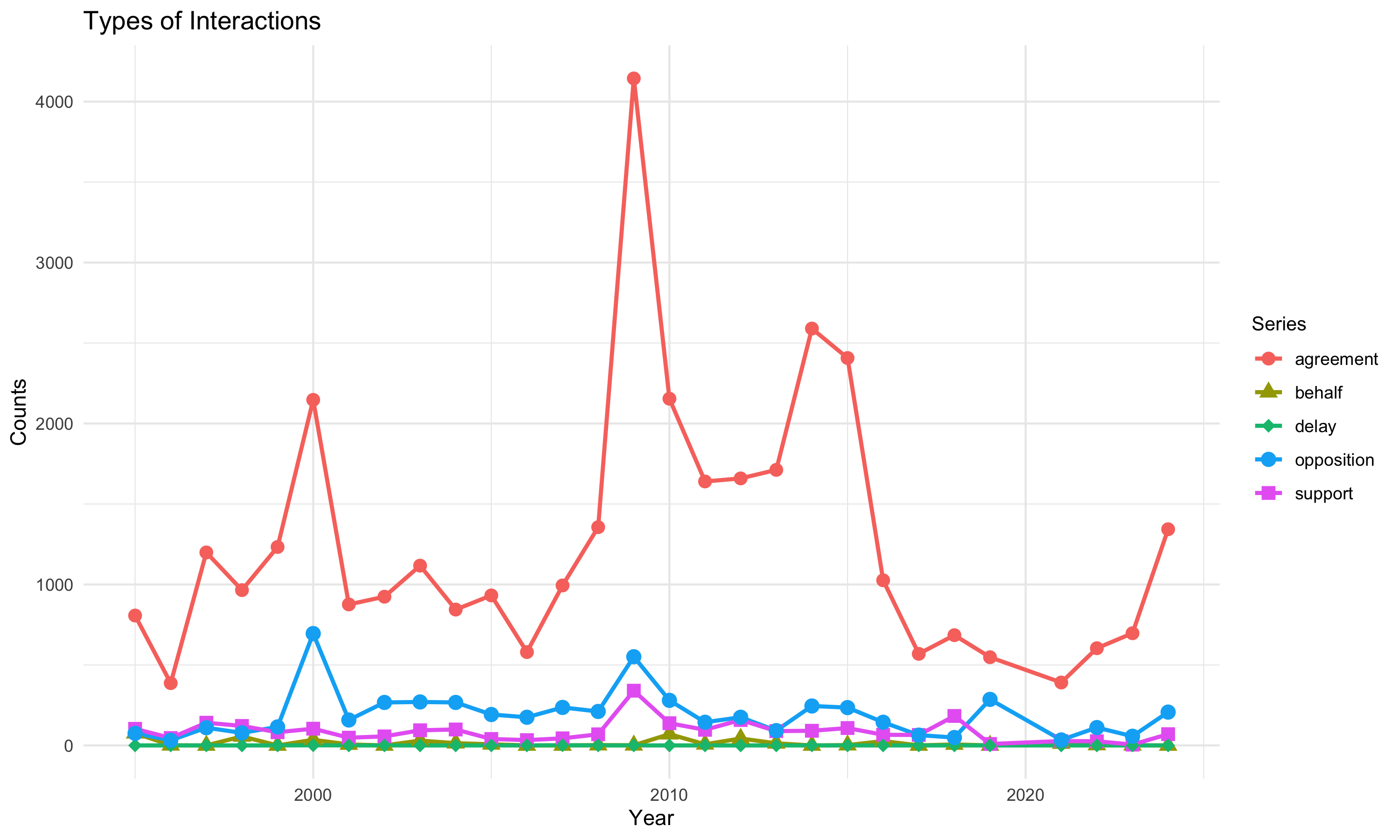} \\
      (a) EALA annotation (1995-2024) \\ 
      \includegraphics[width=0.9\textwidth]{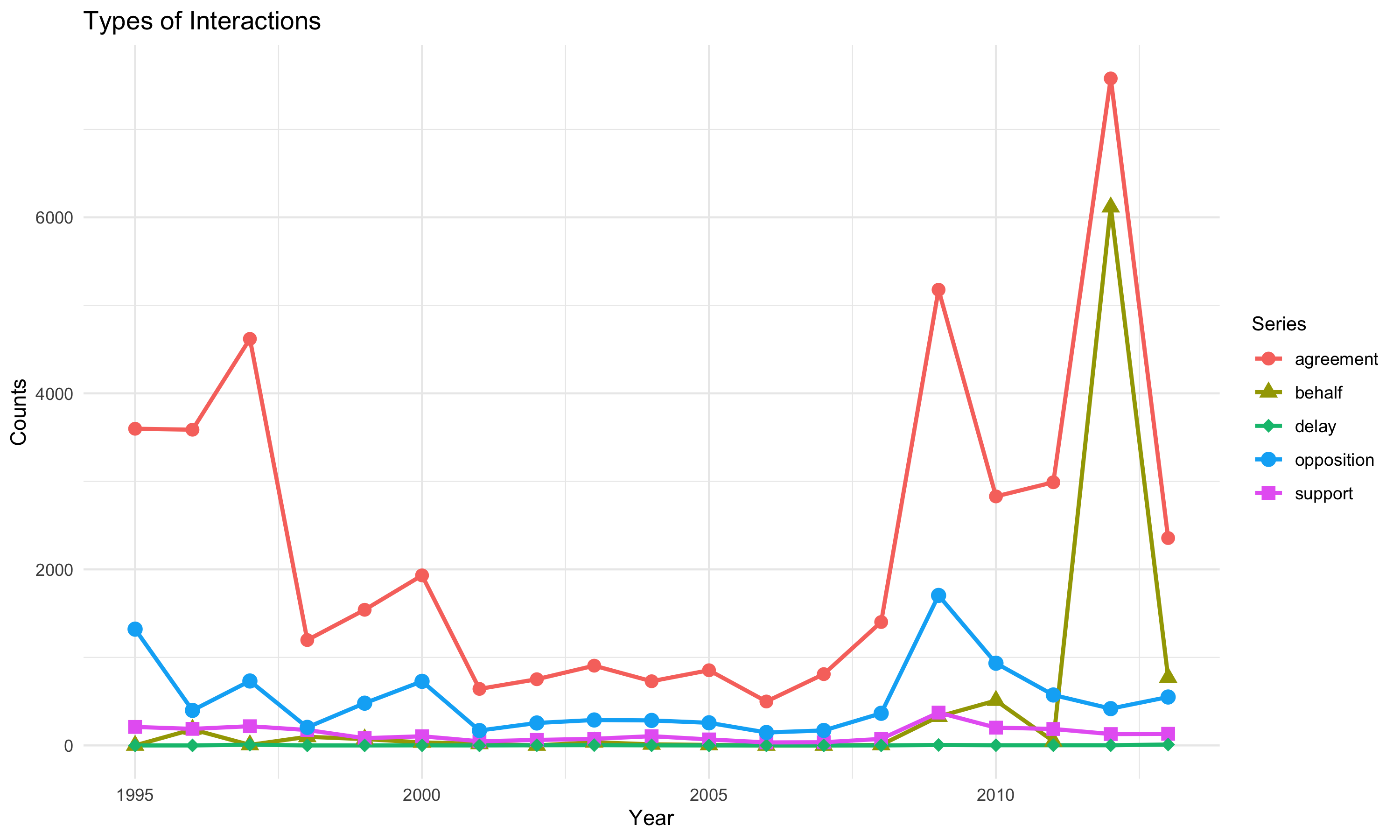} \\   
      (b) Castro annotation (1995-2013) \\
    \end{tabular}
    \caption{Frequencies of the five interaction types.}
    \label{inter}
\end{figure}

\paragraph{Interaction Type} The machine- and human-annotated datasets align closely on the relative frequencies of the five interaction types. As shown in Figure~\ref{inter}, both identify ``agreement'' as the most commonly observed type, followed by ``opposition'', with ``delay'' being the least frequently detected. During the overlapping years, the variation in interaction types is generally similar across the two datasets. However, in 2012, the Castro data shows a notable surge in the ``behalf'' interaction type, making it nearly as frequent as ``agreement''. The very low frequencies observed for the majority of interaction types suggest that it might be more effective to simplify the classification into two broader categories: ``positive'' and ``negative''.

\begin{figure}[ht]
    \centering
    \begin{tabular}{c}
\includegraphics[width=0.9\textwidth]{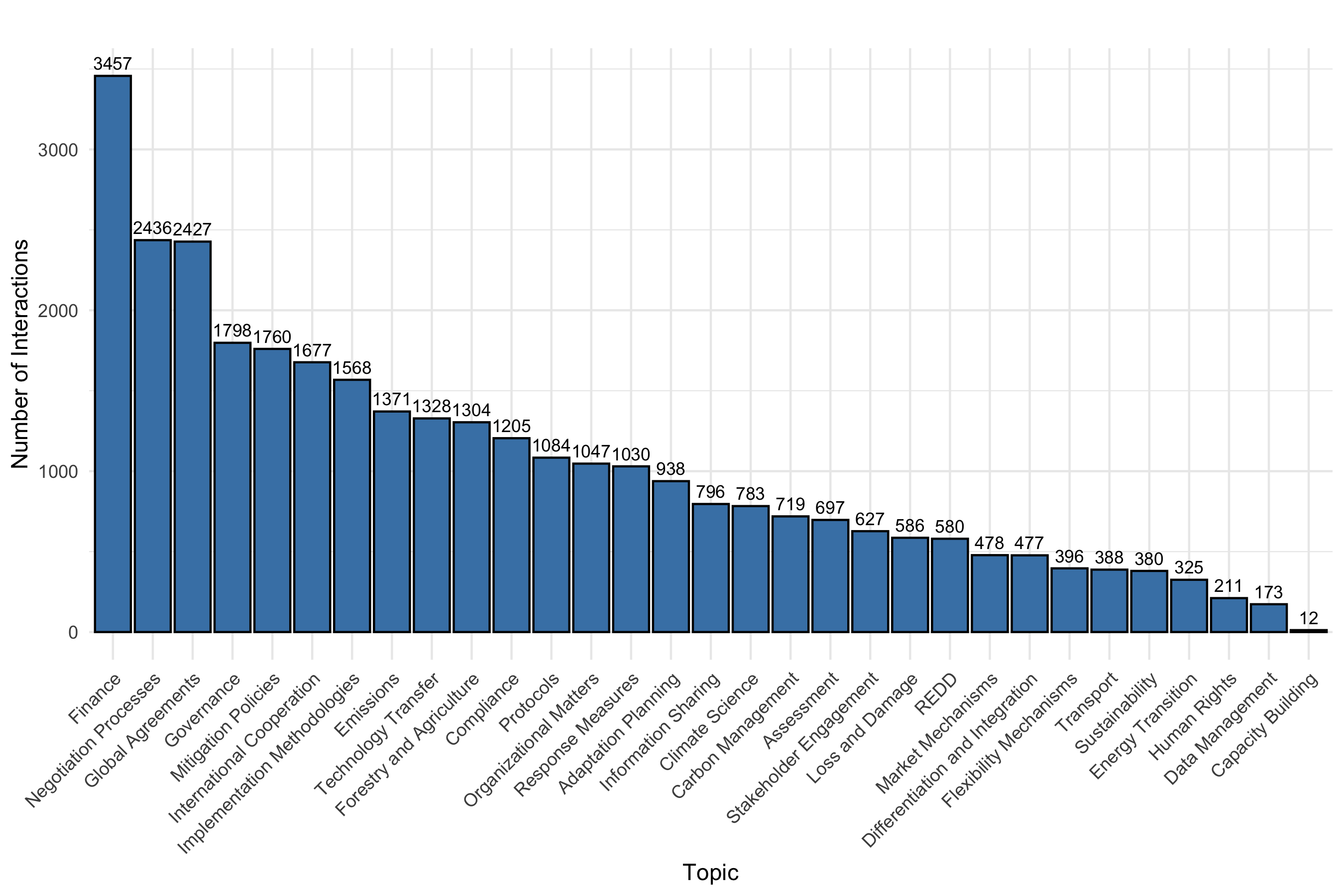} \\
         (a) EALA annotation (1995-2013)\\
      \includegraphics[width=0.9\textwidth]{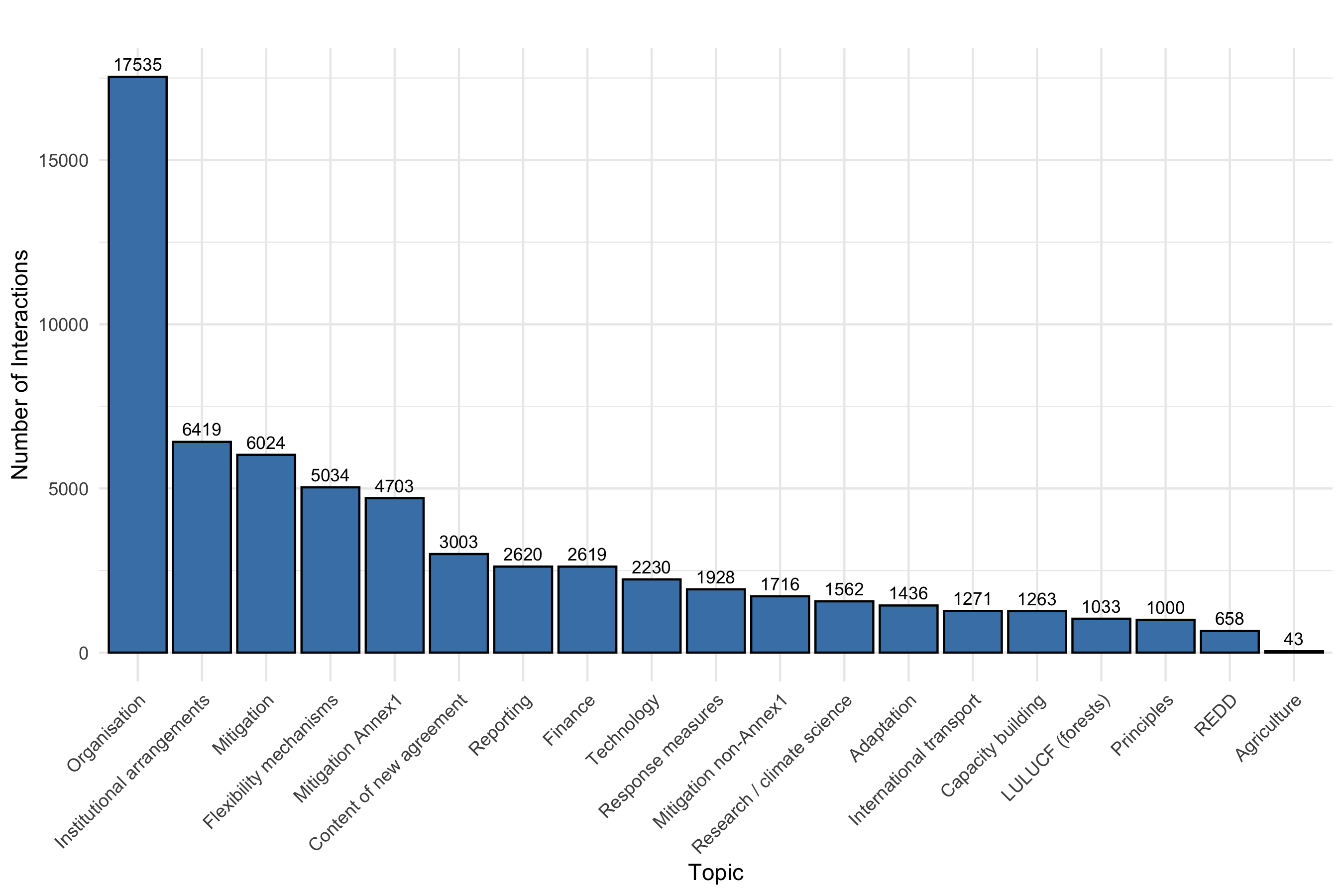} \\
         (b) Castro annotation (1995-2013)\\
    \end{tabular}
    \caption{Distribution of topics.}
    \label{topics.compare}
\end{figure}

\paragraph{Topics} The distribution of topics differs significantly between the EALA and Castro datasets, primarily because we did not adhere to the expert's codebook when instructing LLMs to infer topics of interactions. As shown in Figure~\ref{topics.compare}, EALA identifies substantially more topics than the Castro dataset. Additionally, significant differences are also observed in terms of the relative popularity of these topics. For instance, while EALA identifies ``Finance'' as the most prominent topic in climate negotiations, the Castro dataset ranks ``Organization'' as the top topic, appearing nearly three times more frequently than the second most prominent topic, ``Institutional Arrangement''. Despite some areas of agreement, the EALA data capture a broader range of topics compared to the Castro data. The frequencies of topics annotated by EALA during the full years are reported in Figure~\ref{topics.full}, and Figures~\ref{topics1} and \ref{topics2} in the appendix illustrate the evolution of these topics over the years from 1995 to 2024.

Overall, the EALA-annotated dataset and the human-annotated dataset show substantive alignment in some aspects while diverging in others. It is important to note that the EALA-annotated dataset in the evaluation is entirely machine-generated without any post-processing or manual data cleaning. We strongly caution against the direct use of machine-generated datasets without thorough error correction, which is essential before applying the data to substantive research. Like all datasets, EALA-generated data should be treated as part of a dynamic process involving continuous error reporting and correction. Additionally, the codebook may require periodic updates to incorporate advancements in knowledge and evolving research needs. 

\section{Related Work}


With the recent development of LLMs, a growing number of papers apply LLMs to the annotation of social science tasks, such as sentiment analysis~\citep{ornstein2022train}, hate speech detection~\citep{huang2023chatgpt}, and stance detection~\citep{gambini2024evaluating}. The majority of these studies directly use LLMs with prompting and explore prompt engineering methods~\citep{heseltine2024large, liu2024poliprompt}, while \citet{halterman2024codebook} instruction tune LLMs on annotated data, and \citet{pangakis2024knowledge} generate training labels with LLMs to train supervised machine learning models. 
Among these studies, the most relevant to ours are \citet{xiao2023supporting} and \citet{halterman2024codebook}, which leverage expert-drafted codebooks in annotation. Nevertheless, their focus is on text classification tasks where LLMs already demonstrate competent performance, even in zero-shot settings. In contrast, our paper addresses the complex task of annotating longitudinal network data, aiming to explore the potential of LLMs when assisted with expert knowledge in more challenging application scenarios.

Despite the promising performance of LLMs in automated annotation, there remain issues like the underlying social biases and stereotypes~\citep{ziems2024can} and the non-random prediction errors~\citep{egami2024using}. It is crucial to address these issues to ensure the effective and equitable use of LLMs.

\paragraph{Deductive and Inductive Reasoning of LLMs}
Deductive and inductive reasoning are two important reasoning capabilities. Deductive reasoning concerns the application of explicit rules to derive conclusions, while inductive reasoning involves generalizing patterns from data to make predictions. 

LLMs utilize deductive reasoning for tasks that require rule application. \citet{sanyal2022fairr} tackle logical reasoning tasks by designing rule selection and composition modules, and \citet{feng2024language} internalize the deductive reasoning process through finetuning. The deductive reasoning ability is also vital in low-resource language learning~\citep{tanzerbenchmark, zhang-etal-2024-teaching}. \citet{team2024gemini} find that providing LLMs with the grammar book improves the machine translation performance. 
On the other hand, LLMs exhibit inductive reasoning abilities in in-context learning~\citep{brown2020language,liu2024incomplete}, where models learn patterns from examples provided within the context and apply them to new situations. \citet{wanghypothesis} and \citet{zhu2023large} also show that LLMs are capable of summarizing rules from data, which is an important stage of inductive reasoning. 

In contrast to these works, EALA integrates both deductive and inductive reasoning, showcasing the effectiveness of simultaneously learning from both methodologies, and clearly demonstrating the gains respectively. The effectiveness demonstrated by EALA is particularly noteworthy and significant, given that in social science, deductive reasoning often struggles with the inherent fuzziness of rules, while inductive reasoning must contend with the substantial uncertainty involved in pattern detection.

\section{Conclusion}
This paper introduces EALA, a framework that integrates LLMs with expert knowledge embedded in codebooks and human-annotated historical data to tackle the critical challenge of generating and maintaining timely, high-quality data with complex structures for social science, such as longitudinal network data. By combining deductive learning from rule-based codebooks and inductive learning from annotated examples, EALA provides an effective approach for annotating complex, evolving corpora. Our evaluation, conducted on climate negotiation network data for 30 years, shows that EALA not only outperforms traditional supervised models but also achieves performance comparable to human annotators, making it a promising solution for extending and updating datasets in both data-rich and data-scarce scenarios.

Through careful evaluation, we assess the relative importance of the three key components of the proposed framework: the use of rules retrieved from the codebook as prompts, instruction tuning with human-annotated data, and task decomposition. Based on the climate negotiation data, we find that each of these components significantly enhances the annotation capabilities of LLMs, with instruction tuning emerging as the primary contributor. While LLMs can infer coding rules from annotated data, explicitly providing coding rules that reflect expert knowledge improves the model's ability to capture more nuanced aspects of these rules. Additionally, breaking down complex annotation tasks into subtasks allows LLMs to focus on specific reasoning and inference processes, reducing the risk of hallucinations and improving overall accuracy.

However, EALA has several limitations that require further exploration and refinement. First, LLMs' search windows cannot be excessively wide for complex reasoning tasks, but narrowing the window to a paragraph significantly limits their ability to detect interactions occurring within the same day of negotiation. This limitation might be addressed by adopting LLMs with long context or employing a Retrieval-Augmented Generation (RAG) framework, potentially combined with Chain-of-Thought (CoT) prompting. We believe these approaches are promising avenues for future exploration.
Second, while task decomposition is effective in reducing hallucinations, it does not fully resolve the issue. Here too, integrating RAG with CoT prompting may offer a potential solution. 
Third, LLM-generated predictions are inherently variable, sometimes producing significantly more or fewer annotations than human annotators. This variability, coupled with the limited controllability of the model's reliability and precision, poses challenges in achieving higher accuracy and further optimization. 

Our evaluations suggest that EALA is particularly advantageous for automatically annotating large-scale, rapidly incoming corpora that would otherwise overwhelm human annotators, especially when the task involves inferring complex relationships among a broad set of entities on evolving issues. However, for smaller-scale corpora or when new texts are introduced at a slower pace, human annotation may remain the preferred approach. In addition, for EALA-annotated datasets to be effectively used in downstream political analyses, human intervention to correct errors remains indispensable. In our example, post-prediction rule-based filtering proves to effectively improve the results by reducing inconsistencies and errors, which shows that achieving highly accurate annotations still necessitates manual verification and systematic adjustments to refine the LLM-generated output. 

Despite these challenges, EALA provides a scalable and adaptable tool for generating structured datasets by enabling dynamic label spaces and tackling the inherent difficulties of evolving semantic contexts. This work not only addresses a critical gap in the timely availability of data for political science but also paves the way for broader applications in other fields that rely on longitudinal and relational datasets.


\bibliography{custom}

\appendix
\section{Appendix}
\subsection{The ENB Daily Reports of Negotiations and Human-Annotated UNFCCC Negotiation Longitudinal Network Data}
\begin{figure}[ht]
    \centering
    \begin{tabular}{cc}
      \includegraphics[width=0.47\textwidth]{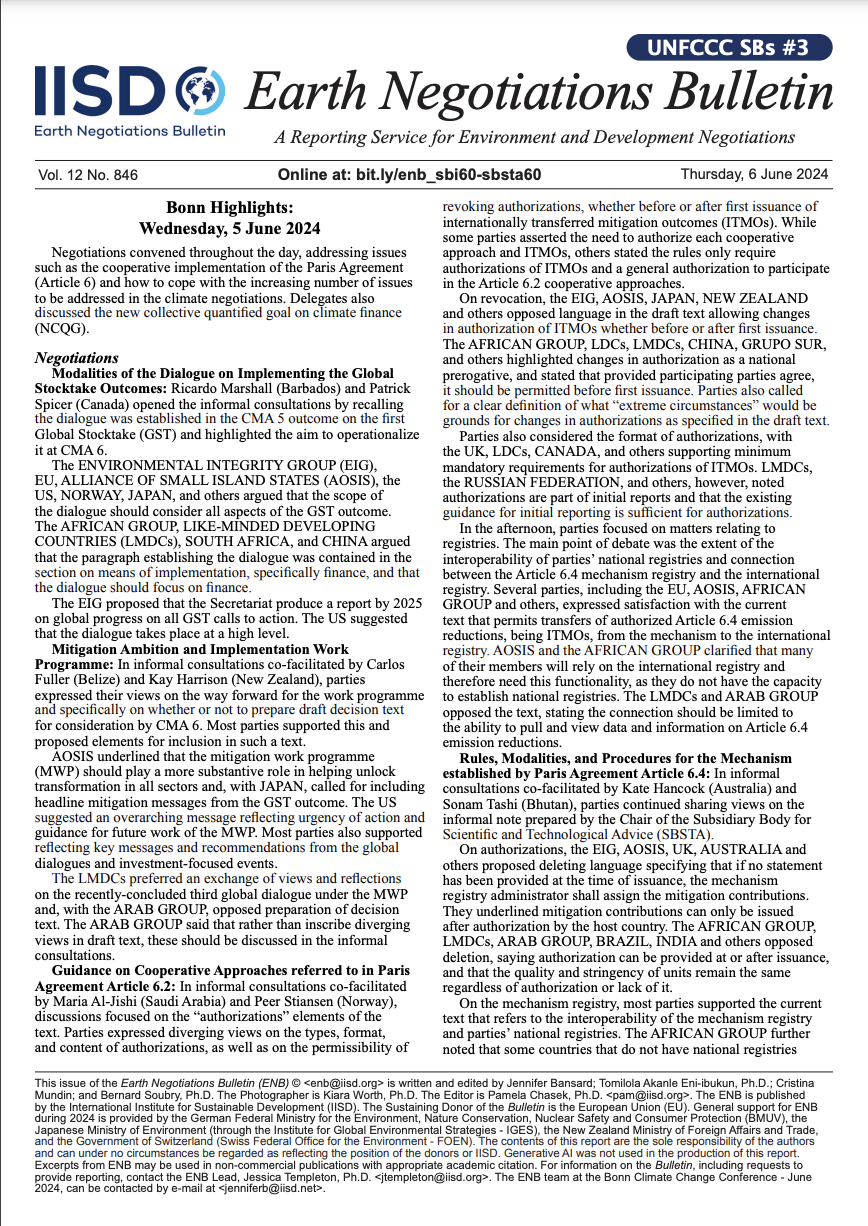} &
      \includegraphics[width=0.47\textwidth]{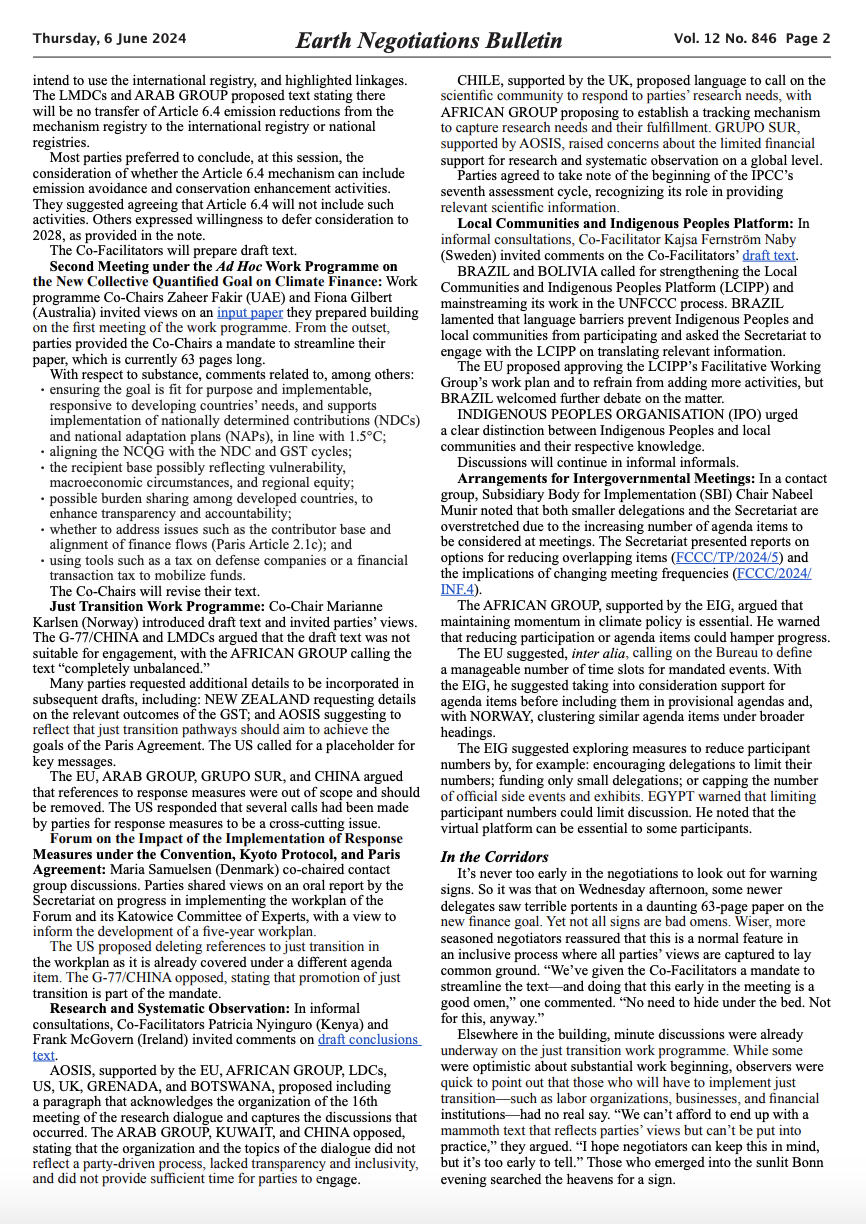} \\
    \end{tabular}
    \caption{An example of daily negotiation report (Bonn Climate Change Conference, for 5 June 2024).}
    \label{ENB}
\end{figure}

\begin{figure}[ht]
    \centering
    \begin{tabular}{cc}
      \includegraphics[width=0.47\textwidth]{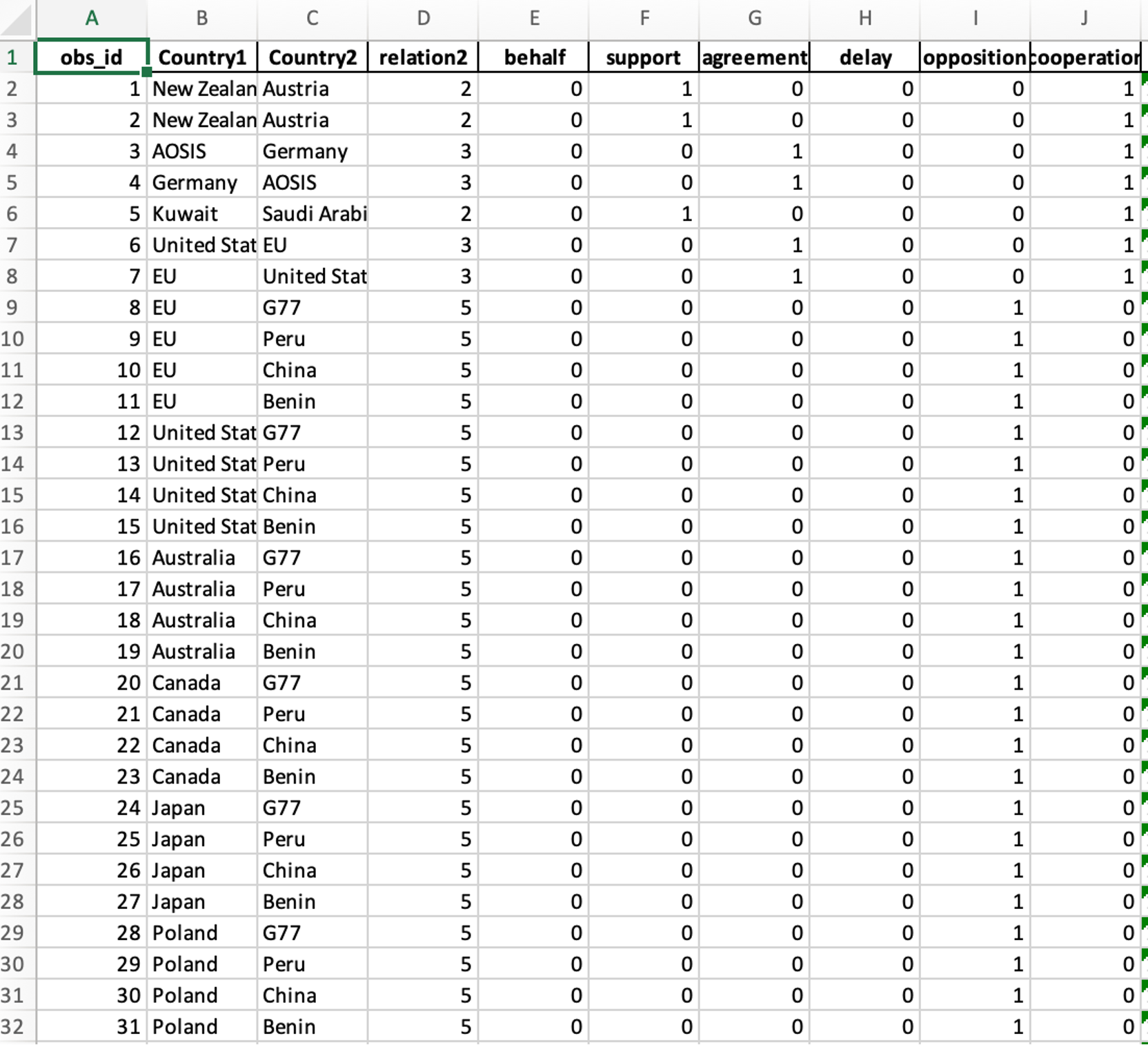} &
      \includegraphics[width=0.47\textwidth]{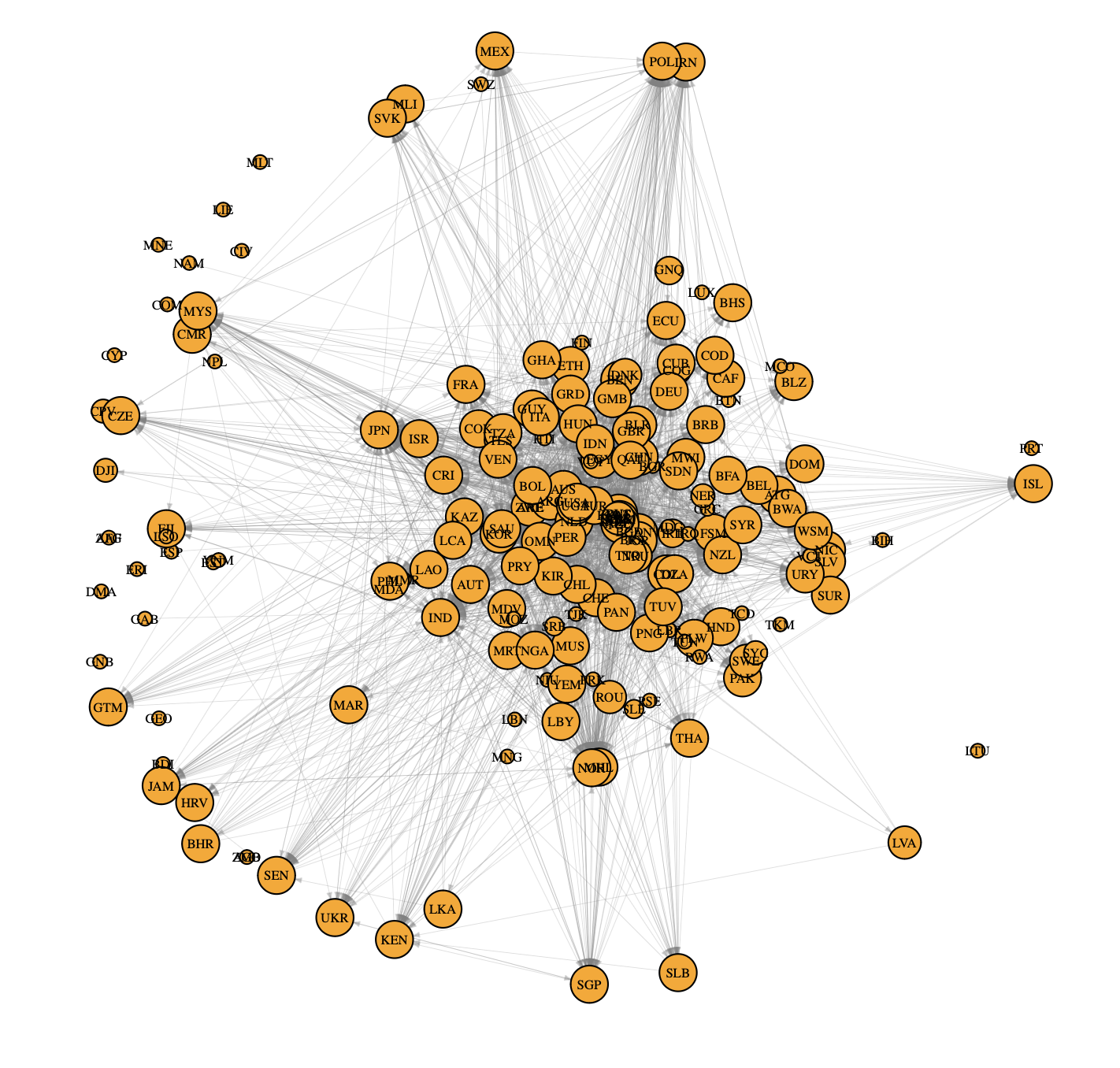} \\
    \end{tabular}
    \caption{Human-annotated longitudinal network data based on experts' codebook.}
    \label{human_data}
\end{figure}

Figure~\ref{ENB} presents an example of ENB daily report of international climate negotiation. Figure~\ref{human_data} showcases what the human-annotated longitudinal network data looks like.

\subsection{Details of Topic Space Construction}
\label{appendix-topic-space}
We utilize GPT-4o to extract topic words from summary reports, and generate names and descriptions for each topic. In the initial stage, we apply the K-means algorithm to group topic words into clusters. The embeddings are generated using Sentence Transformers~\citep{reimers-2019-sentence-bert}, specifically the \texttt{all-MiniLM-L6-v2} version. In the following stages, for each newly extracted topic word, we calculate its similarity to the centers of the existing topic clusters. If the similarity exceeds the smallest similarity between the existing words in the cluster and the center, the word will be added into that cluster. Otherwise, a new topic will be created for the word.

\subsection{Descriptions of the Rules in the Compliance Assessment}
\label{appendix-rules}
To assess how well models comply with rules, we use three rules involving multiple interactions from the codebook of the relation extraction subtask.
\begin{itemize}
    \item \textbf{Bidirectionality}: The relation types agreement and on behalf of are coded bi-directionally. For example, “SAMOA agrees EU” is coded as “Samoa – agreement – EU” and as “EU – agreement – Samoa”.
    \item \textbf{Transitivity}: If several countries (more than 2) agree with each other, then each pair is coded as a new observation, and again in both directions. For example, “AUSTRALIA, NEW ZEALAND, ICELAND and others...” is coded as “Australia – agreement – New Zealand”, “Australia – agreement – Iceland”, “New Zealand agreement – Australia”, “New Zealand – agreement – Iceland”, “Iceland – agreement – Australia”, “Iceland agreement – New Zealand”.
    \item \textbf{Derivation}: If several countries oppose or support another one, then not only the opposition or the support is coded, but also the agreement between all countries that are supporting / opposing.
\end{itemize}

\subsection{The EALA-Annotated Longitudinal Network Data}
\label{llmData}
\begin{figure}[ht]
    \centering
    \begin{tabular}{c}
             \includegraphics[width=0.9\textwidth]{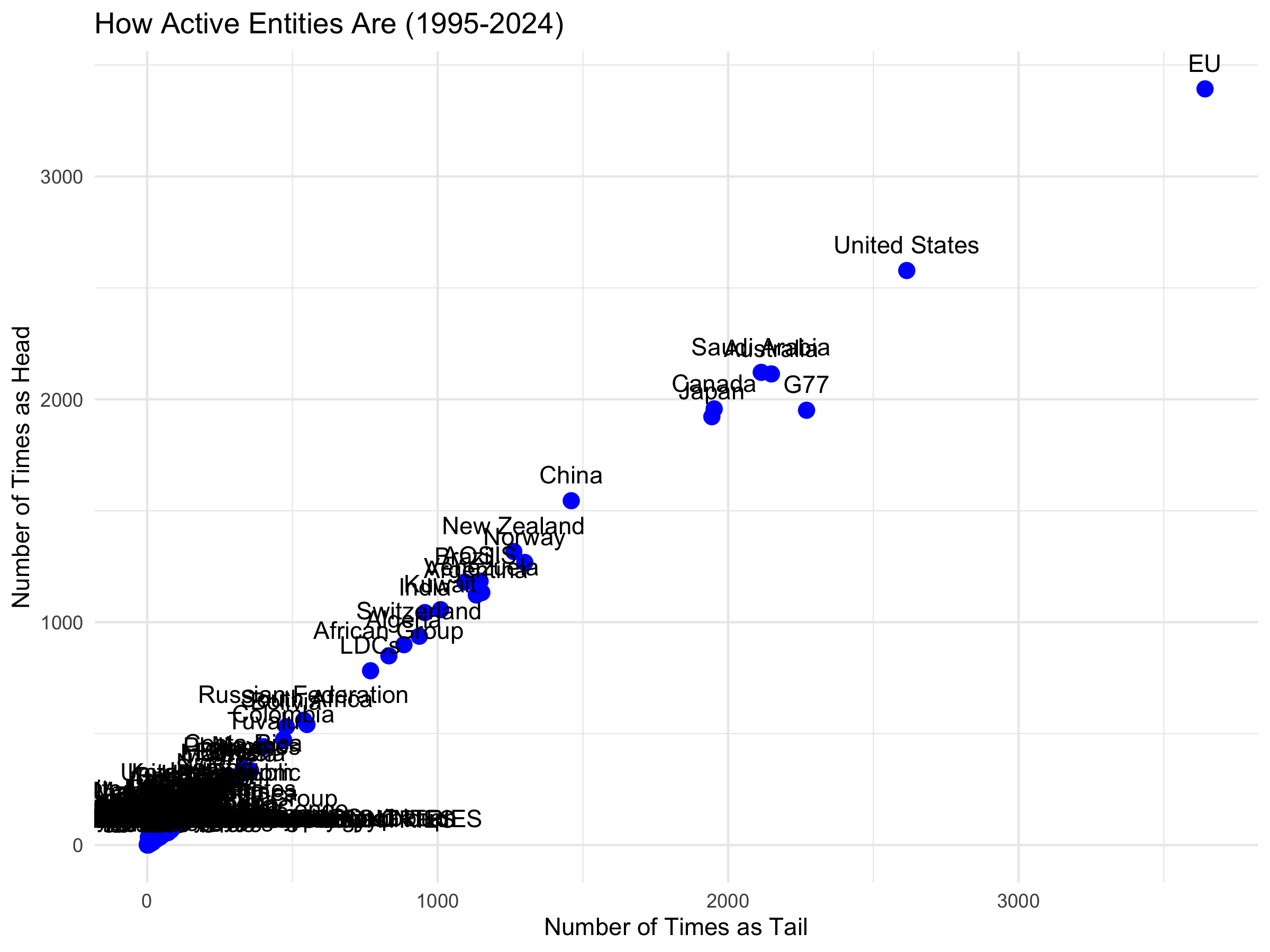} \\
               (c) EALA annotation (1995-2024)
    \end{tabular}
    \caption{Degree of activity of negotiation entities.}
    \label{active2}
\end{figure}

\begin{figure}[ht]
    \centering
    \begin{tabular}{c}
      \includegraphics[width=0.9\textwidth]{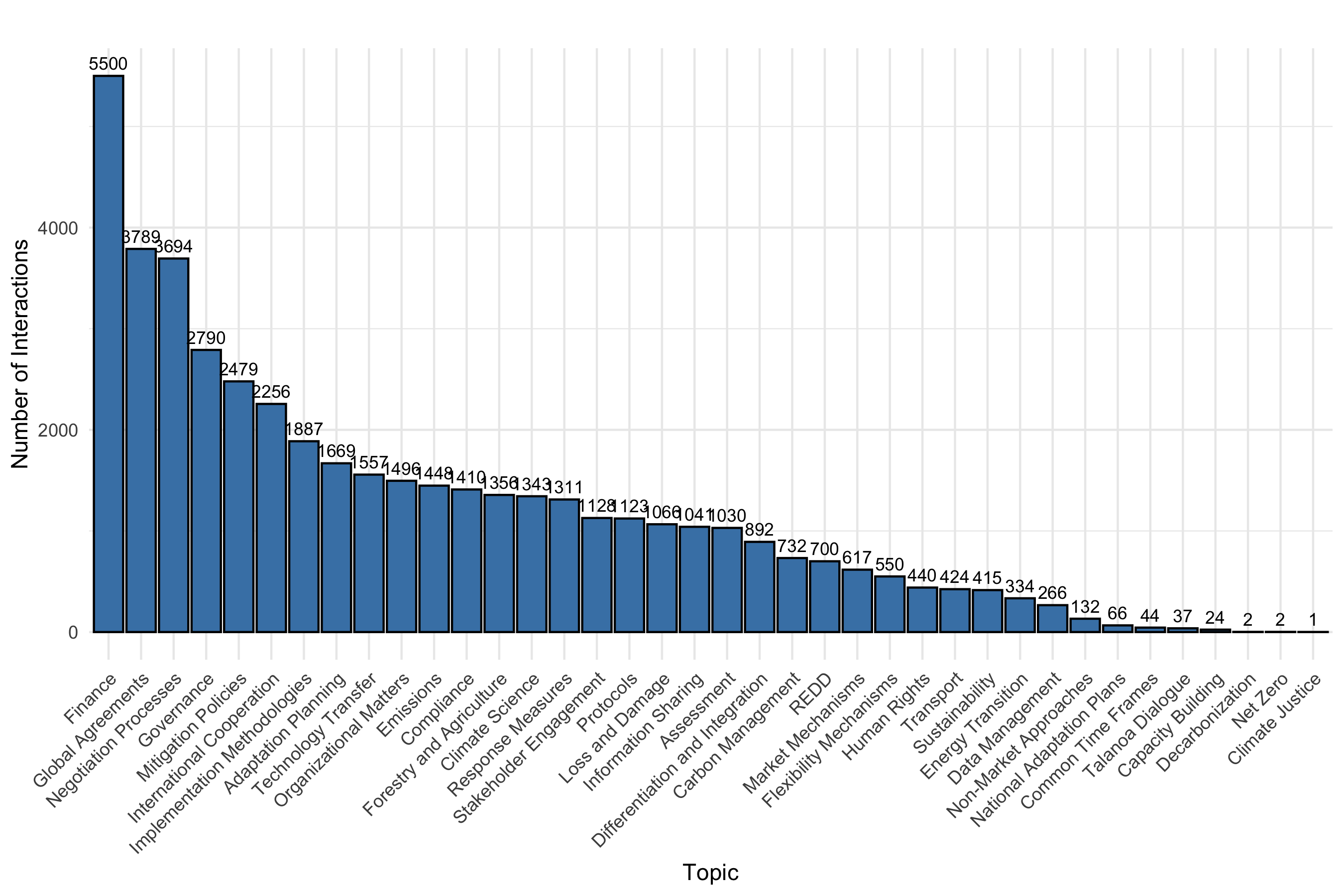} \\
    \end{tabular}
    \caption{Distribution of topics (1995-2024).}
    \label{topics.full}
\end{figure}

\begin{figure}[ht]
    \centering
    \begin{tabular}{c}
      \includegraphics[width=0.8\textwidth]{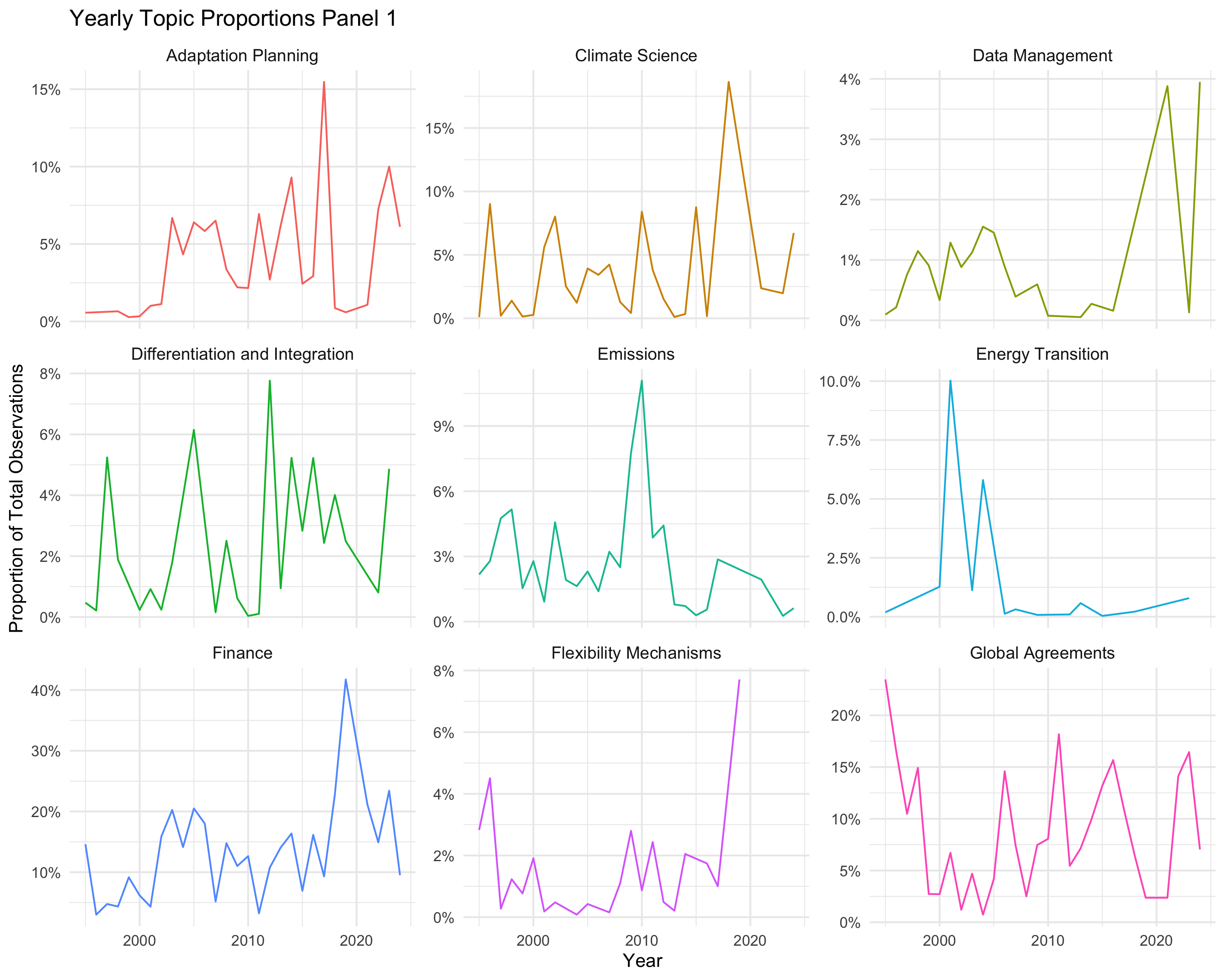} \\
     \includegraphics[width=0.8\textwidth]
      {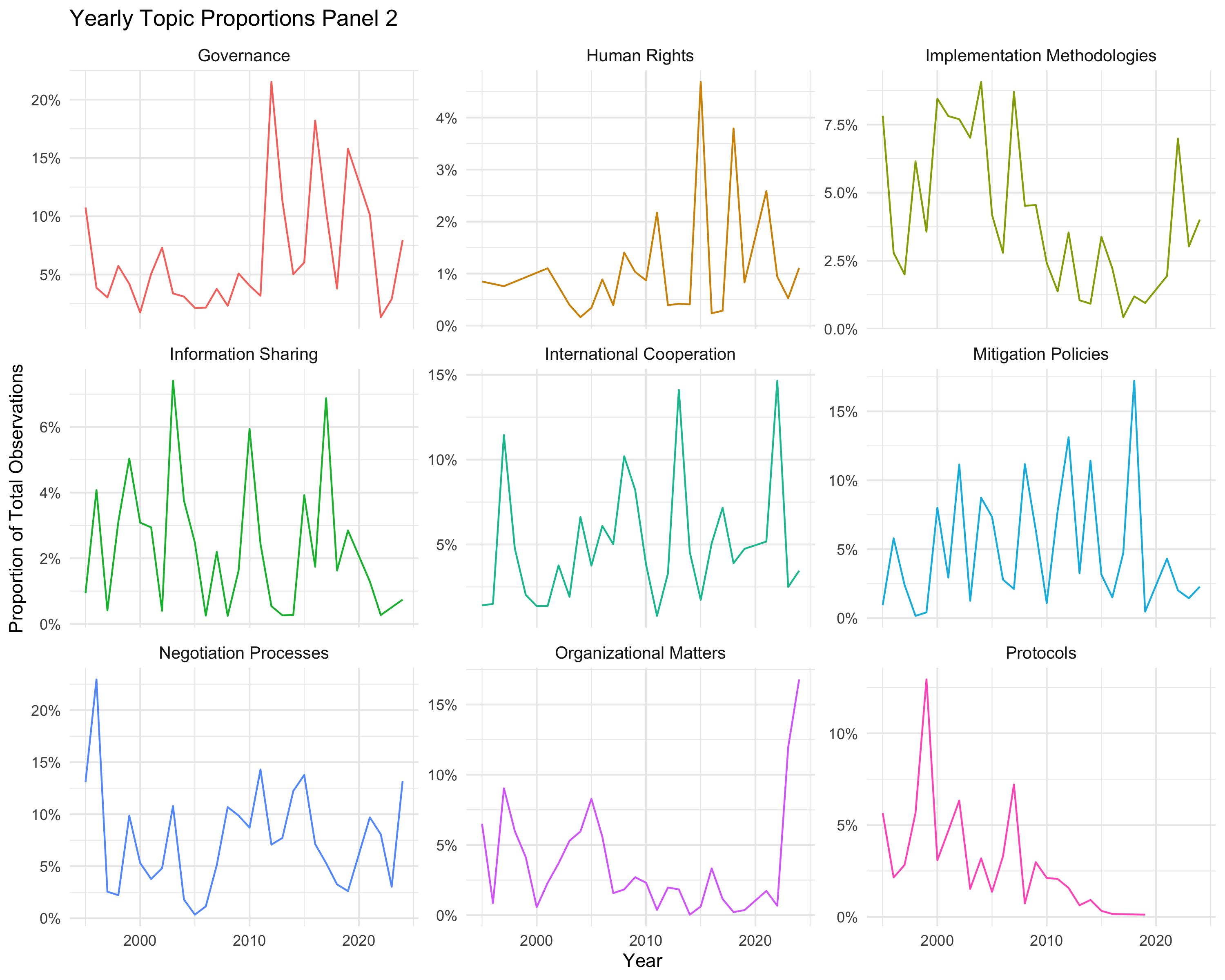} \\
    \end{tabular}
    \caption{The frequencies of topics annotated by EALA (1995-2024).}
    \label{topics1}
\end{figure}

\begin{figure}[ht]
    \centering
    \begin{tabular}{c}
      \includegraphics[width=0.8\textwidth]{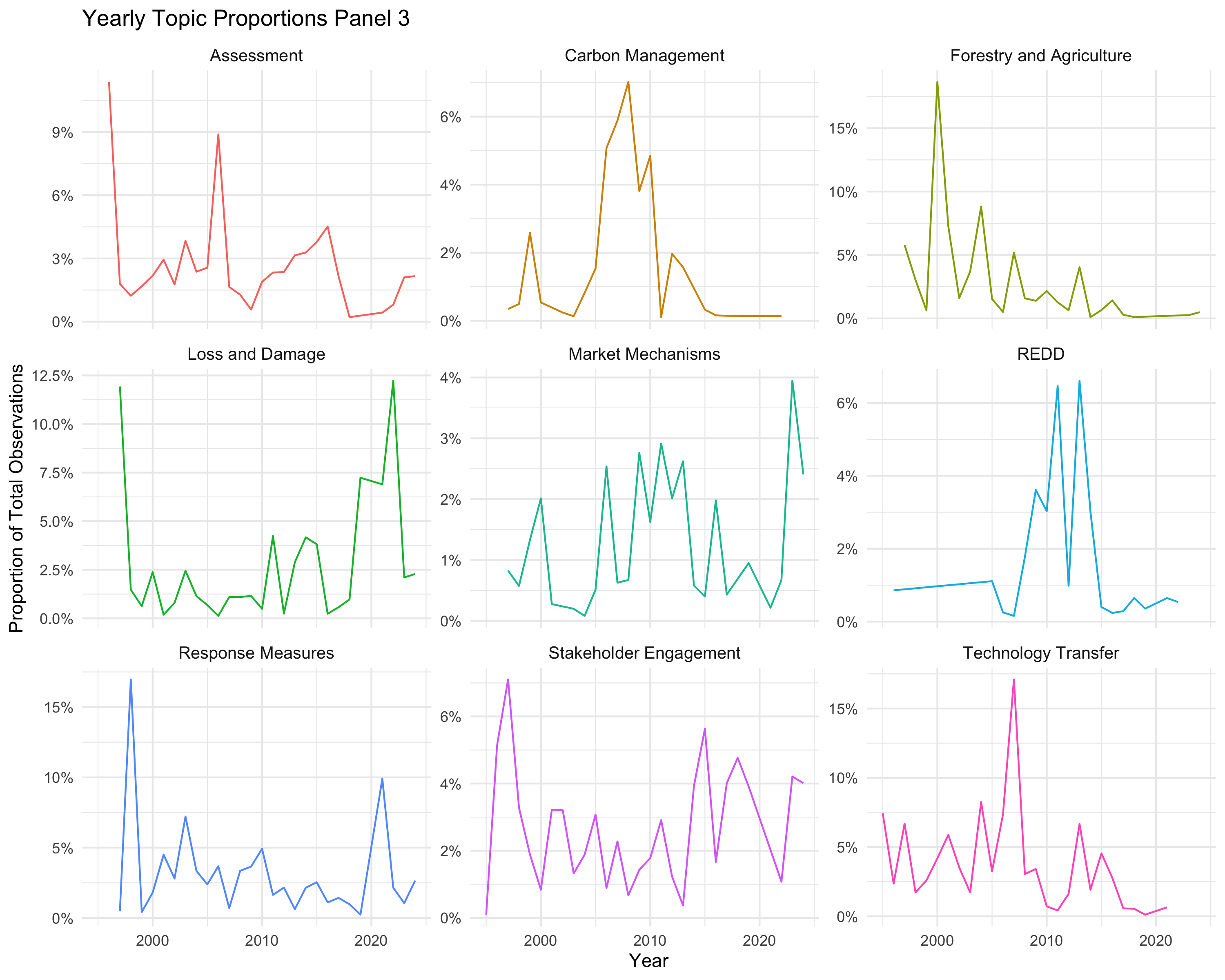} \\
     \includegraphics[width=0.8\textwidth]
      {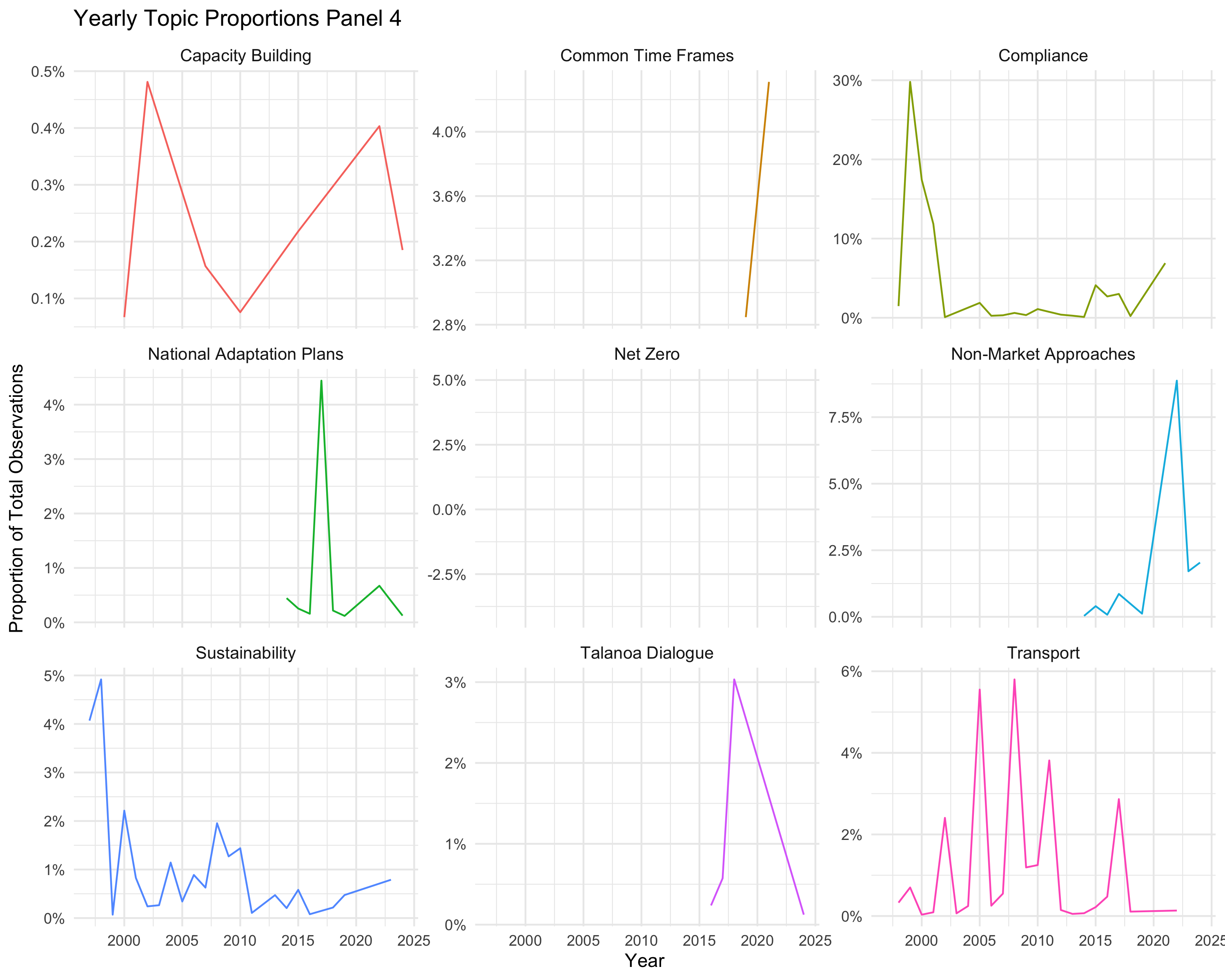} \\
    \end{tabular}
    \caption{The frequencies of topics annotated by EALA (1995-2024). (continued)}
    \label{topics2}
\end{figure}

Figure~\ref{active2} depicts the degree of activity of negotiation entities annotated by EALA during the full period from 1995 to 2024.
Figure~\ref{topics.full} reports the frequencies of topics annotated by EALA during the full period, and Figures~\ref{topics1} and \ref{topics2}  illustrate the evolution of these topics over the years.

\end{document}